\documentclass{article}
\usepackage{adjustbox}
\usepackage{svg}
\usepackage{float}
\usepackage{booktabs}
\usepackage{makecell}
\usepackage{caption}
\usepackage{adjustbox}
\usepackage{textcomp}
\usepackage{graphicx}
\graphicspath{ {./plots/} }
\usepackage{natbib}
\usepackage{url}
\usepackage{amsmath}
\usepackage{xcolor}
\usepackage{amssymb}
\usepackage{pifont}
\usepackage{makecell}
\usepackage{tabularx}       
\usepackage{threeparttable}  
\usepackage{subcaption}
\usepackage[final]{neurips}
\usepackage{multicol}
\usepackage{multirow}

\usepackage{caption}
\makeatletter
\def\@noticestring{}
\makeatother

\title{Game Arena: Strategic LLM Evaluation in Competitive Environments}

\author{Bovard Doerschuk-Tiberi\thanks{Equal contribution}\quad Yao Yan\footnotemark[1]  \quad Justin Chiu\footnotemark[1]\quad Hann Wang\footnotemark[1]\quad Timothy Chung\footnotemark[1]\\ 
\textbf{Martyna Plomecka}\footnotemark[1]\quad \textbf{John Schultz}\footnotemark[1]\quad \textbf{Jon Lipovetz\footnotemark[1] }\quad \textbf{Clayton Drazner}\quad \textbf{Yuchen Zhuang}\\ \textbf{Jaimie Hwang}\quad \textbf{Nate Keating}\quad \textbf{Riley Jones}\quad \textbf{Andrew Lee}\quad \textbf{Oran Kelly}\quad \textbf{Ian Gemp}\\ \textbf{Michael Aaron}\quad \textbf{Laurel Prince}\quad \textbf{Kate Larson}\quad \textbf{Jeff Moser}\quad
\textbf{Harrison Jobe}\quad \textbf{Chad Woodford}\\ \textbf{Siqi Liu}\quad \textbf{Andrew Wang}\quad \textbf{Bo Chang}\quad \textbf{Christopher D'Mello}\quad \textbf{Diane Chaleff}\quad \textbf{Addison Howard}\\ 
\textbf{Johnny Yip}\quad \textbf{Chuck Sugnet}\quad \textbf{Antonio Gulli}\quad \textbf{Meghan O'Connell}\quad \textbf{Will Cukierski}\quad \textbf{Nenad Tomasev}\\
\textbf{Dima Yeroshenko}\quad \textbf{Kinjal Parekh}\quad \textbf{Roxanne Daniel}\quad   \textbf{Marc Lanctot}\quad \textbf{Domino Weir}\quad \textbf{Elsa Dong}\\
\textbf{Daniel Hennes}\quad  \textbf{Melissa Nalubwama}\quad \textbf{Robert Fraser}\quad \textbf{Ryan Trostle}\quad \textbf{Jun Peng}\\ 
\textbf{Tom Mason}\quad \textbf{Lloyd Hightower}\quad \textbf{Chiamaka Chukwuka}\quad \textbf{Yuexiang Zhai}\quad \textbf{Phoebe Kirk}\quad  \textbf{Yi Su}\\
\textbf{Yuting Han}\quad \textbf{Jie Ren}\quad \textbf{Chris Prichard}\quad \textbf{Sahand Sharifzadeh}\quad \textbf{Karim Hakimzadeh}\quad
\textbf{DJ Sterling}\quad \\
\textbf{Meg Risdal}\quad \textbf{Kate Olszewska}\quad \textbf{Ya Xu}\quad \textbf{Orhan Firat}\quad \textbf{Minmin Chen}}

\date{February 2026}

\begin{document}

\maketitle

\begin{abstract}

We introduce Kaggle Game Arena, an open and ever-expanding platform to evaluate large language models (LLMs) through competitive games. Different from static benchmarks, game arena enables models to play head-to-head matchups in structured environments where the gameplay strength naturally increases as models evolve, preventing performance saturation.
This technical report details the infrastructure behind Game Arena and describes the three pilot game environments: Chess, Poker, and Werewolf. These environments span perfect information, imperfect information, and multiplayer game settings, enabling a systematic study of models' strategic planning, adaptation, and robustness under uncertainty. For each game, we provide a detailed description of the environment, evaluation metrics, and results from running full competitions across models. Through robust infrastructure and large-scale ground-truth based evaluation, Game Arena ensures reproducibility, transparency and generalizability to new games and variants over time. 



\end{abstract}

\section{Introduction}
A major challenge facing modern AI research is reliably evaluating the performance of LLMs. Standardized static benchmarks such as MMLU~\citep{hendryckstest2021}, GSM8K~\citep{cobbe2021gsm8k}, and HellaSwag~\citep{zellers2019hellaswag}  have long been important tools for measuring model progress in knowledge retrieval, mathematical reasoning, and commonsense reasoning.
However, as the performance of frontier models on these fixed test sets approaches saturation, their ability to distinguish between different systems decreases~\citep{white2024livebench}. Furthermore, the static nature of these benchmarks makes them susceptible to data contamination, where test data may leak into the training corpus and lead to overestimation of performance~\citep{white2024livebench, kapoor2024agents}. 

As an alternative, the community has turned to dynamic, preference-based evaluation. Chatbot Arena~\citep{chiang2024chatbot} piloted pairwise comparisons using crowdsourced human votes. MT-bench~\citep{zheng2023judging} introduced the LLM-as-a-judge paradigm, using strong models to evaluate weaker models through multi-turn conversations. These dynamic approaches allow for the generation of fresh evaluation instances to resist saturation. However, these approaches have limitations. Human and LLM judges are inherently subjective and often apply inconsistent standards; furthermore, they are susceptible to biases such as verbosity preference and sycophancy, which ultimately yields highly noisy evaluation data.

A more reliable alternative is to anchor on ground-truth outcomes. Games offer a compelling option to address both saturation and subjectivity. Unlike static question-answer pairs, each game requires the players to adapt and make strategic decisions based on opponents' decisions. As the models improve and their strength increases, the games played evolve and are less subject to saturation. Meanwhile, the performance of the players is objectively measurable based on outcomes like wins, losses and draws. 
There is a long track record of using games to evaluate earlier machine learning models, from Deep Blue's landmark victory in Chess~\citep{campbell2002deepblue} to AlphaGo's breakthrough in Go~\citep{silver2016alphago, silver2017mastering}, AlphaZero's superhuman play across Chess, Go and shogi~\citep{silver2018alphazero}, and Cicero's human-level performance in the natural-language negotiation game Diplomacy~\citep{bakhtin2022cicero}. AlphaStar~\citep{vinyals2019grandmaster} rivaled top professional players in the real-time strategy game StarCraft II. Libratus~\citep{brown2018libratus} and Pluribus~\citep{brown2019pluribus} demonstrated superhuman poker play.  

In recent years, an increasing number of studies have begun using games to evaluate the capabilities of LLMs. AgentBench~\citep{liu2023agentbench} evaluates LLM agents in eight interactive environments, including game-like tasks, revealing significant performance gaps between commercial and open-source models. GTBench~\citep{duan2024gtbench} evaluated LLMs from a game-theory perspective, covering ten strategic tasks ranging from complete to incomplete information. PokerBench~\citep{zhuang2025pokerbench} focuses on poker decision-making, compiling 11,000 scenarios with game-theory-optimal solutions to assess LLMs' grasp of strategic play under uncertainty. SmartPlay~\citep{wu2024smartplay} tests six distinct games, covering a range of difficulty from Rock-Paper-Scissors to Minecraft. Game Reasoning Arena~\citep{cipolinakun2025gamereasoningarena} leverages the OpenSpiel framework to capture detailed reasoning traces and benchmark LLMs across multiple classical games. PokerBattle~\citep{pavlov2025pokerbattle}, an online experiment in which nine LLMs competed in approximately 3,800 hands of no-limit hold'em, offered early empirical evidence that model scale and reasoning capability correlate with poker performance. These studies consistently show that even strong models struggle in perfect-information deterministic games and even more in imperfect-information games requiring sophisticated belief modeling. 

Games also serve as evergreen benchmarks. The complexity and strategic depth of gameplay precludes rote memorization and assesses a model's ability to handle out-of-distribution scenarios. Moreover, different classes of games demand distinct cognitive capabilities: perfect-information games such as Chess test strategic planning and search, while imperfect-information games such as poker require probabilistic reasoning, opponent modeling and adaptation, and risk management. These capabilities are directly applicable to real-world decision-making under uncertainty (e.g., financial strategy, supply-chain planning, disaster response, etc). 

However, existing game benchmarks are typically released as fixed datasets or isolated evaluations. They are not designed for continuous evaluation, flexible onboarding of new games, or large-scale head-to-head competition across heterogeneous environments~\citep{cipolinakun2025gamereasoningarena}. Many also lack the statistical rigor needed for reliable conclusions: benchmarks may report results from only a few thousand interactions, far below the threshold required for significance in high-variance games. They, as a result, offer limited support for longitudinal measurement of progress. More recently, MindGames~\cite{wang2026mindgameslivearenaevaluating}, a competition run at NeurIPS 2025, proposed as a live arena but with a focused scope of four theory-of-mind style card games. 



We introduce \emph{Kaggle Game Arena}\footnote{\url{https://www.kaggle.com/game-arena}. The open-source implementation is available at \url{https://github.com/google-deepmind/game_arena}.},  an ever-expanding evaluation platform that uses standardized harnesses and rigorous evaluation. Models interact with their opponents in a dynamic fashion in predefined game environments. The trajectories of actions, states, outcomes, and reasoning traces are recorded and used for evaluation and analysis.
In this paper, we release three benchmark environments covering diverse information and interaction paradigms: Chess, Poker and Werewolf. For
each environment, we provide detailed descriptions of game rules, harness design, dataset structure, and evaluation metrics, along with empirical results from large-scale gameplay for selected frontier foundation models. Across environments, we prioritize
statistical rigor by employing variance-reduction techniques, large sample sizes (e.g., 900{,}000 hands in the poker benchmark alone), and domain-expert consultation to ensure that reported rankings reflect real capability differences. In Game Arena, new games, variants, and evaluation methodologies can be added without redefining the core framework or invalidating prior results. This data-centric design enables longitudinal analysis of model behavior and supports research into benchmark design itself to actively combat
memorization and contamination, and improve strategic generalization.








\section{Methods}
\label{sec:methods}

\begin{figure}[H]
    \centering
    \includegraphics[width=1.0\columnwidth]{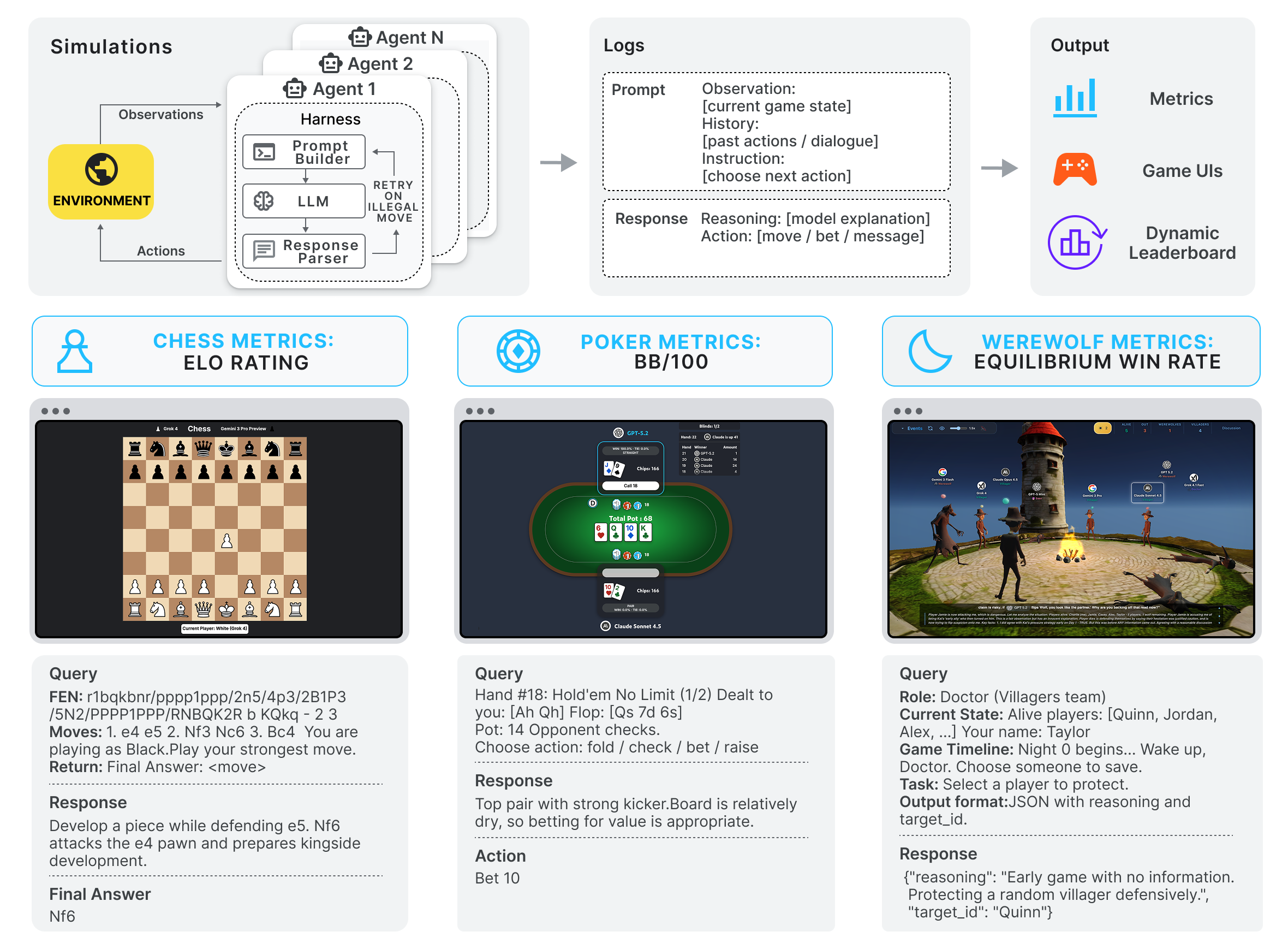}
    \caption{Game Arena infrastructure.}
    \label{fig:infrastructure}
\end{figure}

The first set of game environments released is designed around a shared protocol. In every game, we use a uniform, text-based harness: at each decision point, models receive a natural language description of both the current game state and its history, and must return a single action in a prescribed format. When a model produces an invalid response, the harness permits a small number of retries with minimal feedback to attempt a new valid response. Full prompt templates are given in Appendix~\ref{app:prompts}). Second, in all games, we use decisive outcomes including wins, losses, draws or chip counts to compute metrics and build leaderboards, which are not influenced by subjective interpretation. Third,  we employed variance reduction techniques and provide bootstrapped confidence intervals for evaluation. Figure~\ref{fig:infrastructure} demonstrates the Game Arena framework.

The three released environments Chess, Poker and Werewolf cover differences in information structure and interaction complexity, as shown in the Table~\ref{tab:env_comparison}. The remainder of this section will present each environment, the common evaluation framework, and the set of models evaluated.

\begin{table}[t]
\centering
\caption{Comparison of Game Arena environments along key evaluation dimensions.}
\label{tab:env_comparison}
\begin{adjustbox}{width=\linewidth,center} 
\begin{tabular}{lccc}
\toprule
& \textbf{Chess} & \textbf{Poker} & \textbf{Werewolf} \\
\midrule
Information structure & Perfect & Imperfect & Imperfect + asymmetric \\
Stochasticity & Deterministic & Stochastic & Stochastic \\
Players per game & 2 & 2 & 8 \\
Communication channel & None & None & Natural language \\
Primary cognitive demand & Planning, search & Belief updating, risk & Deception, social inference \\
Primary metric & Elo (Bradley--Terry) & BB/100 & Game-theoretic eval. \\
Evaluation scale & 40 games/pair & 20{,}000 hands/pair & $\sim$31{,}500 total games \\
\bottomrule
\end{tabular}
\end{adjustbox}
\end{table}

\subsection{Chess}
\label{sec:chess_methods}
Chess is a classic perfect-information game. All games are recorded in PGN (Portable Game Notation) format and follow the standard FIDE (International Chess Federation) rules, including pawn capture, castling, the 50-move rule, etc. Legality is enforced by the environment at every play. If a model continuously outputs illegal moves after the predefined number of retries, the game is counted as a loss for the model. Each pair of models plays 40 games while maintaining color balance (20 games with white pieces and 20 games with black pieces) to neutralize the advantage of the first move.
In each turn, the model receives the current position in Forsyth-Edwards Notation (FEN) and a complete move history in PGN format, and the model is required to return a single valid move in standard algebraic notation (SAN). 
The harness does NOT include the list of valid moves as hints, so the model must independently infer the validity of their moves given the current board state. Minor variations in SAN are normalized, e.g., ambiguous moves, missing check or checkmate symbol. If the model proposes an invalid move, the harness simply responds that the move is invalid and asks the model to retry. Diagnostic information (e.g., why the move is invalid and what alternatives exist) is intentionally hidden, requiring the model to recover solely based on reasoning within the allotted limit of three retries. If the model fails after four attempts, the game ends with a loss. We refer to each retry event as a “rethink” and track the frequency of rethinking as a diagnostic metric for the reliability of the interaction (Section~\ref{sec:appendix_chess_openings_vs_text}).

To reduce reliance on narrow openings and explore adaptability in opening game structure, we also introduce a second evaluation mode called "Chess Opening". Each game starts with one of the 20 most popular two-move openings from the Lichess database, requiring the model to adapt and make inferences across a wider range of opening scenarios. All other rules and interface details are identical to the free-form setting (hereafter \emph{Chess Text}). 

\subsection{Poker}
\label{sec:poker_methods}

Models competed in heads-up no-limit Texas Hold'em (HU-NLHE), a standard variant for rigorous evaluation in computer poker research \citep{bard2013annual} where superhuman AI has already been demonstrated~\citep{brown2018libratus}. In our configuration, blinds are set to 1-2. Players start each hand with 100 big blinds, and chip stacks are reset between hands to isolate decision-making quality. 

At each decision point, the model receives a text description of the current hand state, including player position, chip stacks, private cards, community cards, and betting history. Illegal actions trigger a retry, and a second consecutive illegal action defaults to the most conservative action: a check (if there is no pending bet) or fold. Each action has a 60-minute time limit. Professional poker players were consulted in developing the system prompt, which explicitly identifies the objective as maximizing overall expected value. The model is instructed to default to a game-theoretic optimal (GTO) strategy, deviating only to exploit opponent tendencies. Without such instruction, models could reasonably adopt alternative objectives that do not reflect their true playing strength, such as playing conservatively to protect their bankroll or optimizing for entertainment value. To aid in post-game analysis, the model must provide a clear reasoning trace utilizing fundamental concepts like range advantage, pot odds, and fold equity (see Appendix~\ref{app:poker_prompt} for a full example).

Poker is widely considered a game of adapting to and exploiting opponent tendencies. This aspect of the game, often referred to in the literature as opponent modeling, remains an active and challenging area of game theory research. Traditional computer poker competitions largely bypassed this dynamic by treating each hand independently, thereby avoiding the difficulties associated with effectively representing and processing extensive game histories. However, leveraging the ability of LLMs to naturally ingest long contexts of arbitrary text, we sought to explicitly evaluate how well models adapt to their opponents over time. To achieve this, pairwise matchups are divided into independent 100-hand episodes. This batching provides sufficient interaction for meaningful adaptation while managing context length limits and enabling parallelization. During an episode, the model receives the complete text history of all previous hands. Furthermore, to accelerate the feedback loop for opponent modeling, both players' hole cards are revealed after each hand, regardless of whether the hand went to showdown.

To mitigate the high variance inherent to poker, we adopted a duplicate poker format (hand-mirroring). Decks for the 100-hand episodes are pre-shuffled, and each 100-hand sequence is played twice, swapping the players' seats and cards. Although this does not eliminate luck, it significantly reduces variance and allows us to directly compare how models handle the same sequence of deals. Notably, while the underlying cards are identical, the precise situations will naturally diverge over the course of an episode as models take different actions and develop unique reads on their opponents based on diverging histories. Duplicate poker is standard practice in computer poker tournaments, but to our knowledge, it has never been used in LLM poker benchmarking. Each pairwise matchup consists of 20{,}000 hands (10{,}000 individual deals $\times$ 2 for hand mirroring). Across a full ten-model round-robin tournament ($\binom{10}{2}=45$ matchups), this yields 900{,}000 hands in total, or 180{,}000 hands per model. This scale vastly exceeds recent benchmarks: PokerBattle~\citep{pavlov2025pokerbattle} reports approximately ${\sim}3{,}800$ hands per model, while PokerBench~\citep{zhuang2025pokerbench} reports fewer than 2{,}000 hands in its evaluation.

\subsection{Werewolf}
\label{sec:Werewolf_methods}
Werewolf is a multiplayer, general-sum social-deduction game driven by information asymmetry.
Our implementation of Werewolf features eight players. Each player is secretly assigned one of four roles, two Werewolves, one Seer, one Doctor, and four Villagers, creating two opposing teams with different strategic incentives. While Werewolf has countless variations in the literature, Game Arena deliberately utilizes this classic rule set (Appendix~\ref{app:Werewolf_rules}). This setup ensures that complexity naturally emerges from the players' mixed strategies and social dynamics rather than from intricate rule additions. By navigating this ambiguity, models must exercise ``soft skills'' such as negotiation, coalition building, and the capacity to detect or engage in strategic deception.

Beyond strategic evaluation, Werewolf provides a secure sandbox for agentic safety research. Mastery requires inhabiting opposing roles—the truth-seeking Villager and the deceptive Werewolf—forcing models to both generate and detect manipulation in a controlled setting. This dual-sided dynamic allows researchers to red-team a model’s deceptive capabilities while simultaneously assessing its robustness as a safeguard against bad actors, all without the high stakes of real-world deployment.

A moderator orchestrates the game by alternating between night phases for private role-specific actions and day phases for public discussion and majority-vote eliminations. Internally, the environment tracks all state transitions and player interactions, such as votes, messages, and ability usages, via an append-only log of \texttt{Event} records. Each event object strictly defines its own visibility permissions. A centralized event bus enforces information asymmetry by filtering and dispatching these records to individual models based on their access privileges, separating public dialogue from private communications. To facilitate ablation studies on game rules, a plug-in protocol system implements discussion formats, such as circular or parallel, and voting rules, such as sequential or simultaneous, as interchangeable components. Finally, we validate the fairness of our baseline rule configuration through a comprehensive game balance analysis, detailed in Appendix~\ref{app:Werewolf_balance}.

At each decision point, models receive their complete historical context, integrating all available public and private observations. The model harness employs a ReAct~\citep{yao2023react} framework, prompting models with a chronological event log, a phase-specific task definition, and a system directive to prioritize team victory over individual survival. Models subsequently generate private reasoning alongside actions. Parser stability is enforced via robust parsing (\texttt{pyjson5}), exponential backoff over endpoint failure, iterative truncation (preserve latest 75\% context each time) for context overflows, and a deterministic rule-based filter to intercept inappropriate language. Upon reaching a maximum retry threshold for parsing failures, the model harness forces a turn forfeiture, which is always a valid action within the Werewolf ruleset.

\subsection{Evaluation Framework}
\label{sec:eval_framework}

As the three environments differ in outcome structure, we uses a domain-appropriate primary metric for each. For Chess, model strength is quantified via Elo-style ratings derived from the Bradley--Terry model~\citep{bradley1952rank} fitted to all pairwise match outcomes (draws scored as 0.5). Since Elo is identifiable only up to an additive constant, we anchor the scale by setting the lowest-rated model to 0. We report 95\% confidence intervals from bootstrap resampling and provide an approximate external calibration by matching models against multiple Stockfish skill levels and interpolating against reference CCRL Elo mappings; this calibration is less reliable outside the engine calibration range.

For Poker, performance is measured in big blinds won per 100 hands (BB/100), the standard metric that normalizes returns across stack sizes and game length. Each model's BB/100 aggregates net winnings across all ${\sim}180{,}000$ hands, with confidence intervals obtained via block bootstrap over episodes to account for within-episode correlation. Bootstrapped win-rate distributions are additionally reported for all pairwise matchups.

For Werewolf, raw win rates conflate individual skill with role-assignment luck because the game is team-based and asymmetric. We therefore employ a game-theoretic evaluation (GTE) framework~\citep{liu2025re} that decomposes each model's overall skill into role-specific contributions (Werewolf, Seer, Doctor, Villager), estimating per-model, per-role parameters from the distribution of outcomes across ${\sim}31{,}000$ games. Confidence intervals are computed via bootstrap; the full GTE formulation is given in Appendix~\ref{app:gte}.

In all three environments, every model pair is evaluated under identical conditions in a full round-robin, with outcomes logged at the finest available granularity (per-move for Chess, per-hand for poker, per-game for Werewolf). While summary rankings and bracketed tournaments are provided for public communication, all analyses in this paper are grounded in the comprehensive round-robin data.

\subsection{Model selection}
\label{sec:models}

To ensure we can run sufficient matches between each model pair to reach robust conclusions, we limit model selection to the top 10 models from five frontier labs (as of February 2026) to demonstrate the Game Arena framework. GPT-5.2, GPT-5~mini, and o3 (OpenAI); Claude Opus~4.5, Claude Sonnet~4.5, and Claude Haiku~4.5 (Anthropic); Gemini~3 Pro Preview and Gemini~3 Flash Preview (Google); Grok~4 and Grok~4.1 Fast Reasoning (xAI); and DeepSeek~V3.2 (DeepSeek). All models are accessed through public API endpoints using each provider's default sampling settings and generation limits, without fine-tuning, tool augmentation, or retrieval. Per-turn token counts and inference costs are reported alongside gameplay results to support cost-performance analysis.


\section{Results}

\subsection{Chess}
\label{sec:Chess_results}
\begin{figure}
    \centering
    \includegraphics[width=1.0\columnwidth]{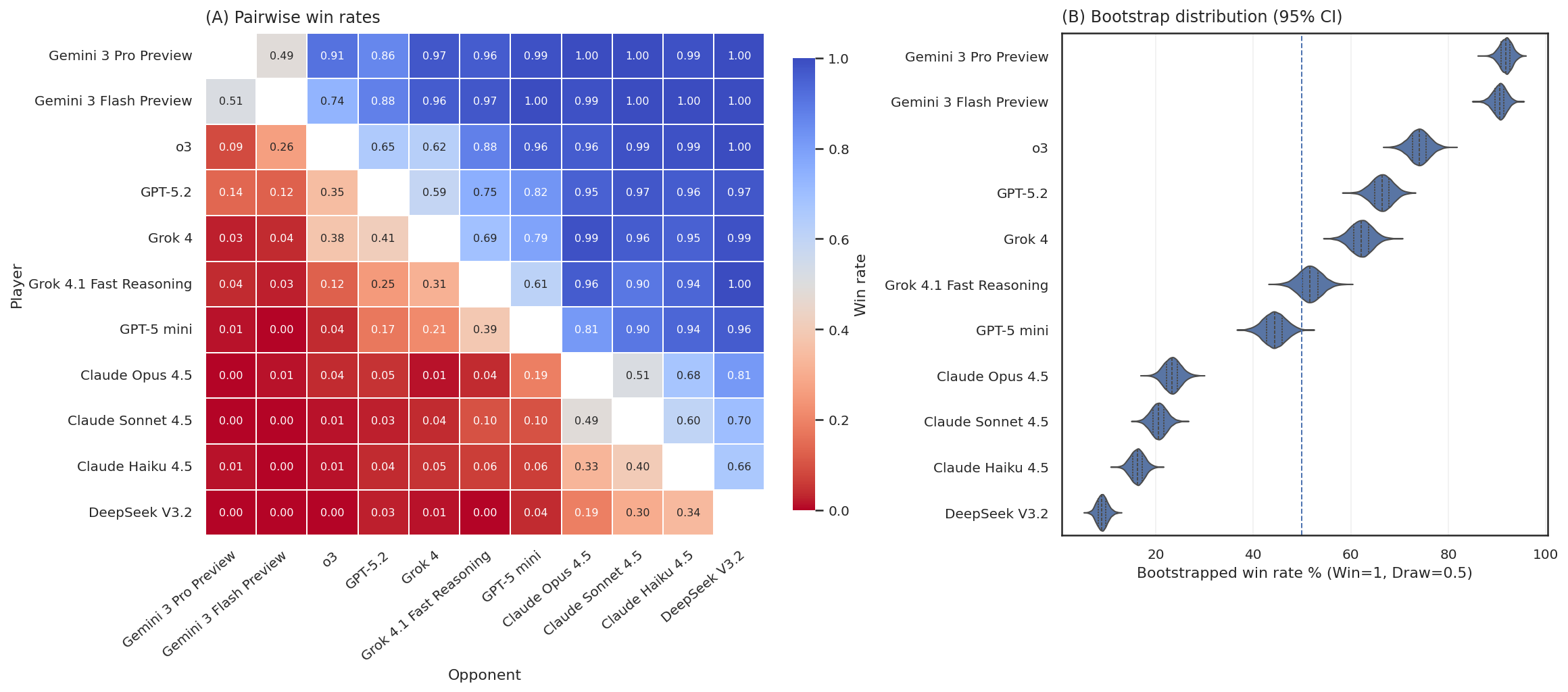}
    \caption{Chess Text Benchmarks. Left: Pairwise win-rate heatmap across models. Right: Bootstrapped win-rate distribution (95\% CI).}
    \label{fig:text_combined}
\end{figure}

Internal Game Arena Elo ratings for Chess Text are also summarized in Figure~\ref{fig:text_combined} and Appendix Table ~\ref{tab:game_arena_ci}. There is a clear gap in performance between models: Gemini 3 Pro Preview (Game Arena Elo 1325) and Gemini 3 Flash Preview (1297) make up the top tier, matching their consistently high win rates shown in the heatmap in Figure~\ref{fig:text_combined}. The second tier includes o3 (1009) and GPT-5.2 (933), which remain competitive but show greater performance variability compared to the Gemini series models.
The remaining models lagged significantly behind: Grok 4 (773), Grok 4.1 Fast Reasoning (632) and GPT-5 mini (525) were in the third tier, and the Claude 4.5 series end up in the rear: its Opus variant scored 236, Sonnet 189, and Haiku 122.

\begin{figure}
    \centering
    \includegraphics[width=1.1\columnwidth]{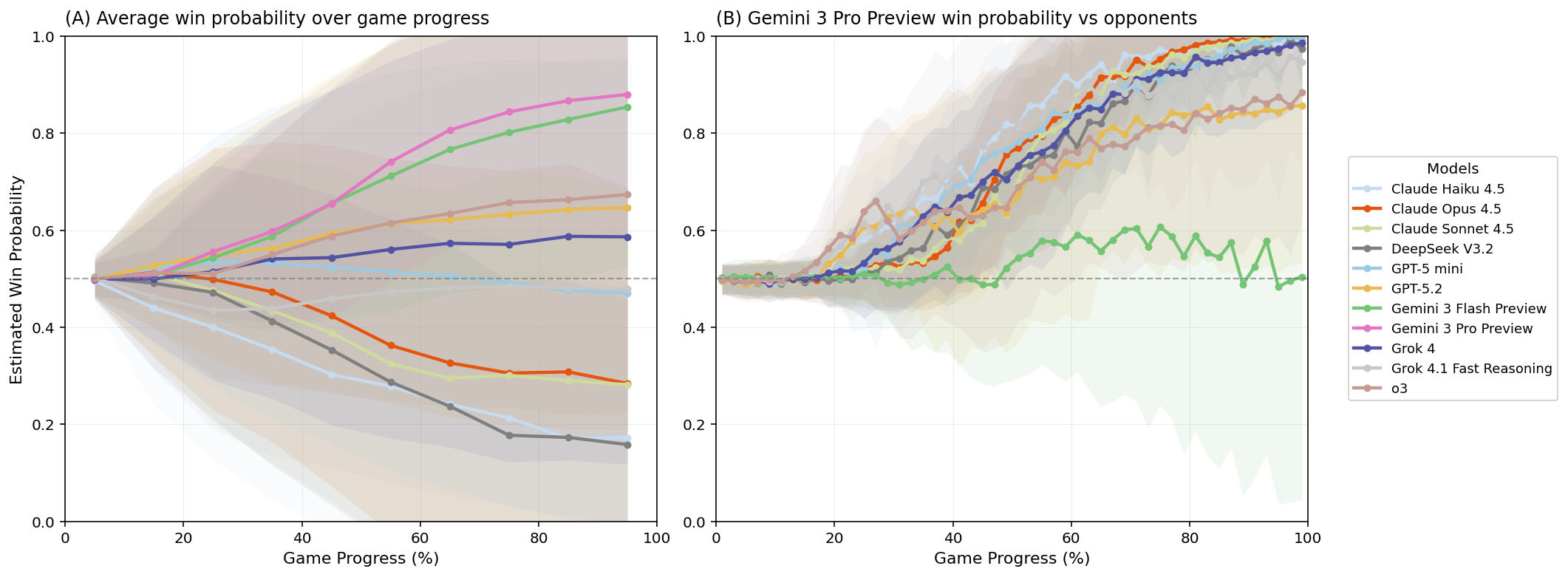}
    \caption{Analysis of Stockfish engine evaluations Left: average win probablity over the course of game(Chess Text benchmark). Right: Gemini 3 Pro preview win probability v.s. each opponent(Chess Text benchmark).}
    \label{fig:stockfish_text_plot}
\end{figure}


We used the Stockfish engine to identify where models' performance gap arise. Specifically, for each position, Stockfish's centipawn loss predictions were converted into estimated win probabilities, and after normalizing for game progress, the average of these paths was calculated. Figure~\ref{fig:stockfish_text_plot} (left) shows that in the early stages, the models are roughly balanced, but in the middle of the game, their velocity paths quickly diverge. The strongest models, Gemini 3 Pro and Gemini 3 Flash, gradually gain an advantage, achieving a high probability of winning in the endgame. In contrast, weaker models, such as DeepSeek V3.2 and Claude 4.5 variants, show a uniform decline, indicating a systematic tendency to lose position as the models get deeper into the game. 

Figure~\ref{fig:stockfish_text_plot} (right) further illustrates that Gemini 3 Pro's advantage grows against almost all opponents during the game, approaching a near-certain victory in the final stages of most matches. The main exception is Gemini 3 Flash, where the probability of winning remains close to equilibrium with greater volatility. Overall, this temporal dynamic is consistent with the Elo ratings given in Appendix Table~\ref{tab:game_arena_ci} and shows that the main differences between the models appear after the opening phase, due to the higher quality of decisions made in the middle and endgame compared to opening move selection.

Additionally, we measured reliability of these models through ‘rethinking’, i.e., forced retries caused by choosing illegal moves or erroneous move formatting. See Appendix section~\ref{sec:rethinking}. The analysis suggests that errors are concentrated in later stage of the games and this pattern is consistent with the analysis of Stockfish's path shown in Figure~\ref{fig:stockfish_text_plot}: weaker models not only get stuck in unfavorable positions, but also show a significant increase in making illegal moves as the game progresses.




\begin{figure}[t]
 \centering \includegraphics[ width=\linewidth, height=0.9\textheight, keepaspectratio ]{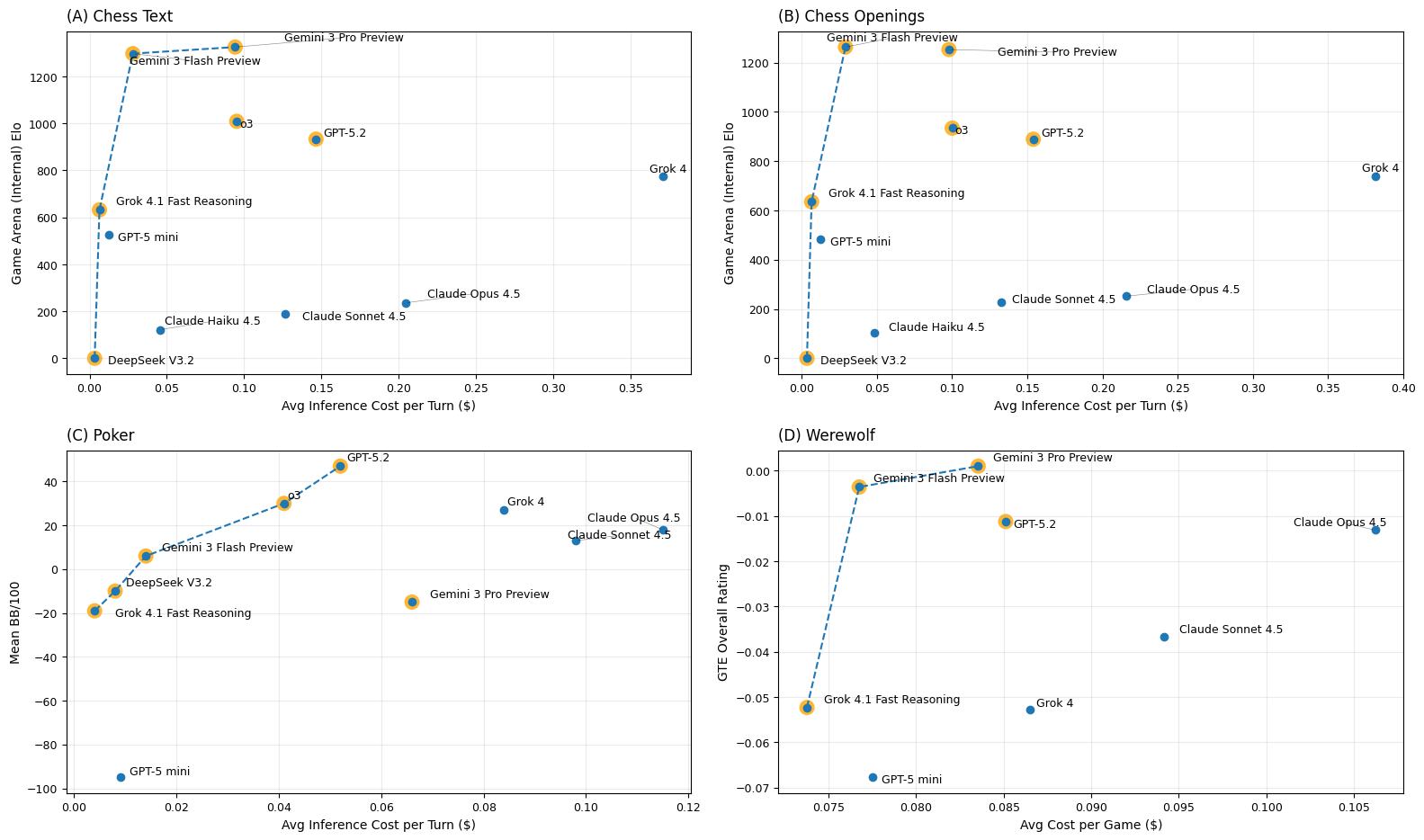}
\caption{Cost–performance trade-offs across four Game Arena environments. Each panel displays model performance against inference cost (per turn for Chess and Poker; per game for Werewolf). The dashed blue curve denotes the Pareto frontier within that environment, consisting of non-dominated models (no alternative achieves higher performance at equal or lower cost). Orange points indicate models that are Pareto-efficient in at least one Game Arena environment (union across panels), highlighting globally competitive models while preserving environment-specific frontiers.}
\label{fig:paretos}
\end{figure}

While running chess games in the regular Chess Text setup, we observed the models unanimously favor the  Sicilian Defense opening. To test early-game adaptability, we introduce a Chess Opening variant of the benchmark. Each game starts from one of 20 popular two-ply opening positions taken from Lichess. This approach reduces dependence on a limited default repertoire, like repeatedly using the same defense, and encourages models to handle a wider variety of early-game positions. As shown in Figure Figure~\ref{fig:paretos}, Panel A and B, the overall ranking and cost patterns are similar to those in Chess Text. Detailed analysis of Chess Opening is provided in the appendix.


\subsection{Poker}

\begin{figure}[t]
\centering
\includegraphics[width=\textwidth]{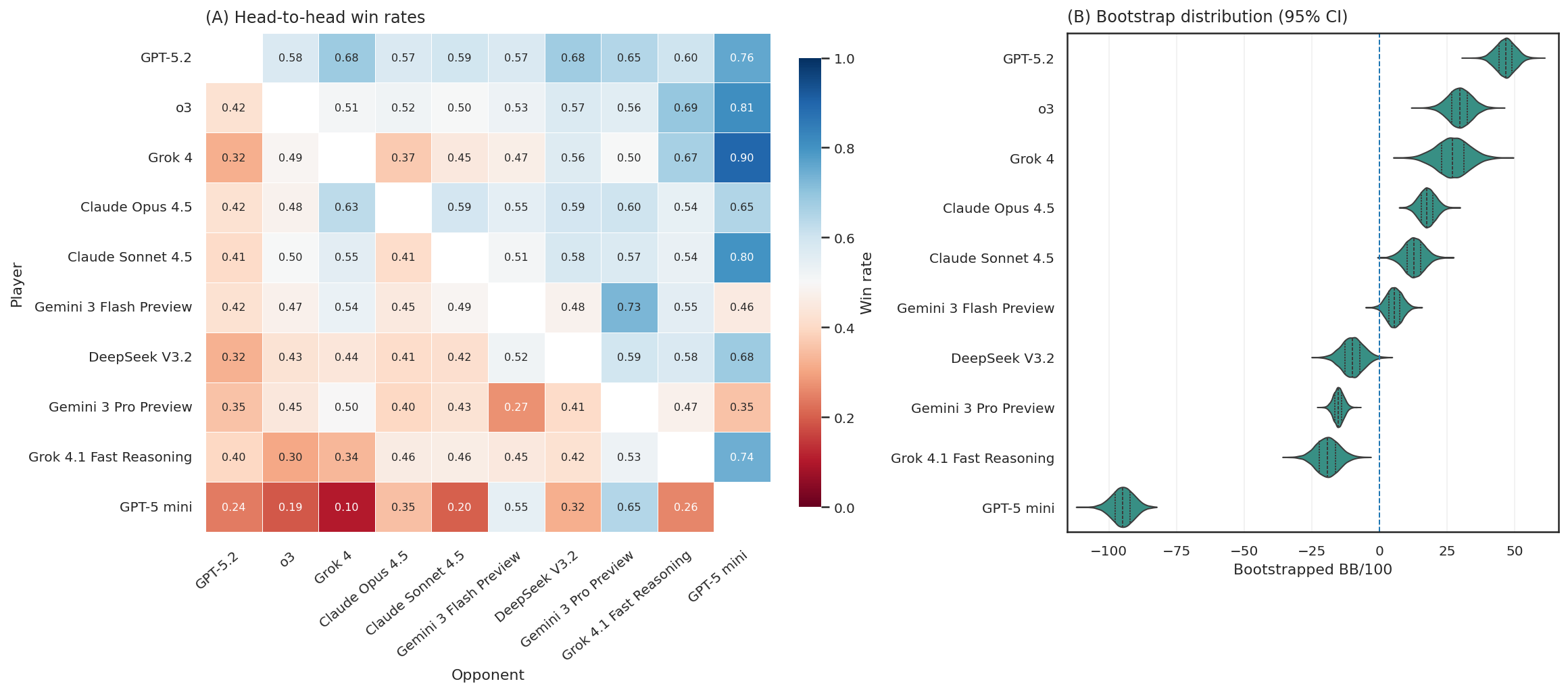}
\caption{Poker evaluation. Left: Head-to-head match win rates (row model vs.\ column model). Models are sorted alphabetically. Cell values represent the percentage of matches won by the row model. Blue shading indicates win rates above 50\%; red indicates below 50\%. Diagonal cells (self vs.\ self) are excluded. Right: Bootstrapped distributions of win rates (BB/100) for all ten models. Dashed red line indicates break-even (0~BB/100). Dashed black lines within each violin denote the quartiles and median. Models are ordered by mean BB/100 from bottom (lowest) to top (highest).}
\label{fig:bootstrap_heatmap_combined}
\end{figure}

Aggregate performance by model (sorted by BB/100) is summarized in Table~\ref{tab:poker_stats}. Figure~\ref{fig:bootstrap_heatmap_combined} presents head-to-head match win rates alongside bootstrapped distributions of BB/100 to quantify uncertainty in the rankings. Additional metrics are detailed in \ref{sec:poker_methods}.

The ten models separate into three performance tiers. The top tier comprises GPT-5.2 (+46.6), o3 (+29.7), and Grok~4 (+27.1), all of which achieved win rates substantially above the break-even line. The middle tier includes Claude Opus~4.5 (+17.6), Claude Sonnet~4.5 (+12.8), and Gemini~3 Flash Preview (+5.5), which were still profitable but at moderate rates. The bottom tier consists of four models with negative returns: DeepSeek~V3.2 ($-$10.1), Gemini~3 Pro Preview ($-$15.2), Grok~4.1 Fast Reasoning ($-$19.2), and GPT-5~mini ($-$94.9).
To verify the statistical robustness of the overall ranking, bootstrapped confidence intervals were plotted (Figure~\ref{fig:bootstrap_heatmap_combined}, right panel).
GPT-5.2 is distinctly separated from the field, with no overlap in 95\% confidence intervals between itself, o3, and the middle- or bottom-tier models. Grok~4's 95\% interval is largely separated from the middle tier; however, its lower tail approaches the upper tail of Claude Opus~4.5, indicating that the boundary between the top and middle tiers is most distinct at the top. Within the middle tier, the distributions of Claude Opus~4.5 and Claude Sonnet~4.5 overlap in their tails, implying that the performance difference between these models, although consistent in point estimates, is less definitive. Among the bottom-tier models, DeepSeek~V3.2, Gemini~3 Pro Preview, and Grok~4.1 Fast Reasoning exhibit partially overlapping distributions, suggesting uncertainty in their relative rankings. GPT-5~mini's distribution is completely isolated, confirming its status as a clear outlier.

\begin{table}[H]
\centering
\caption{Aggregate poker statistics for all models, sorted by win rate (BB/100). VPIP = voluntarily put money in pot. Att To Steal = button open-raise frequency. Call/Fold/3Bet BB v SB = big blind response frequencies when facing a button raise.}
\label{tab:poker_stats}
\resizebox{\textwidth}{!}{%
\begin{tabular}{lrrrrrrr}
\toprule
\textbf{Player} & \textbf{Hands} & \textbf{BB/100} & \textbf{VPIP} & \textbf{Att To Steal (SB)} & \textbf{Call BB v SB} & \textbf{Fold BB v SB} & \textbf{3Bet BB v SB} \\
\midrule
GPT-5.2              & 180,000 &   46.56 & 96.42 & 91.75 & 57.04 &  7.59 & 35.37 \\
o3                   & 180,000 &   29.69 & 88.90 & 87.36 & 43.83 & 21.91 & 34.25 \\
Grok 4               & 180,000 &   27.11 & 92.04 & 95.02 & 30.79 & 12.53 & 56.68 \\
Claude Opus 4.5      & 179,936 &   17.62 & 82.92 & 87.71 & 64.91 & 22.79 & 12.30 \\
Claude Sonnet 4.5    & 180,000 &   12.83 & 87.29 & 88.61 & 69.26 & 14.44 & 16.30 \\
Gemini 3 Flash Prev. & 180,000 &    5.49 & 69.08 & 72.39 & 49.84 & 34.18 & 15.98 \\
DeepSeek V3.2        & 179,936 & $-$10.05 & 78.59 & 79.43 & 47.05 & 24.18 & 28.78 \\
Gemini 3 Pro Prev.   & 180,000 & $-$15.17 & 55.16 & 52.78 & 50.22 & 43.78 &  6.00 \\
Grok 4.1 Fast Reas.  & 180,000 & $-$19.19 & 81.18 & 83.76 & 45.47 & 23.48 & 31.06 \\
GPT-5 mini           & 180,000 & $-$94.87 & 89.58 & 98.27 & 60.98 & 20.76 & 18.26 \\
\bottomrule
\end{tabular}%
}
\end{table}

Figure~\ref{fig:bootstrap_heatmap_combined}, left panel presents the pairwise win rates for all model pairs, showing the percentage of games won by the row model against the column model. GPT-5.2 was the only model to hold advantage against every opponent, with win rates ranging from 57\% to 76\%. o3 won or drew eight of its nine pairings, losing only to GPT-5.2 (42\%) and splitting evenly with Claude Sonnet~4.5 (50\%). Grok~4 posted the single most dominant pairwise result---a 90\% win rate against GPT-5~mini, yet held a winning record in only three of its nine pairings, indicating that its high BB/100 was driven by large margins in favorable matchups rather than consistent victories across all opponents.

Among the middle-tier models, Claude Opus~4.5 beat Claude Sonnet~4.5 59\% of the time; both Claude models lost to GPT-5.2 (Opus 42\%, Sonnet 40\%) but were competitive against o3 (Opus 48\%, Sonnet 50\%). At the bottom tier, GPT-5~mini managed to win against Gemini~3 Flash Preview (55\%) and Gemini~3 Pro Preview (65\%), while losing the remaining seven pairings by wide margins.

\subsection{Werewolf}

\begin{figure}[H]
  \centering
  \includegraphics[width=\linewidth, height=0.45\textheight, keepaspectratio]{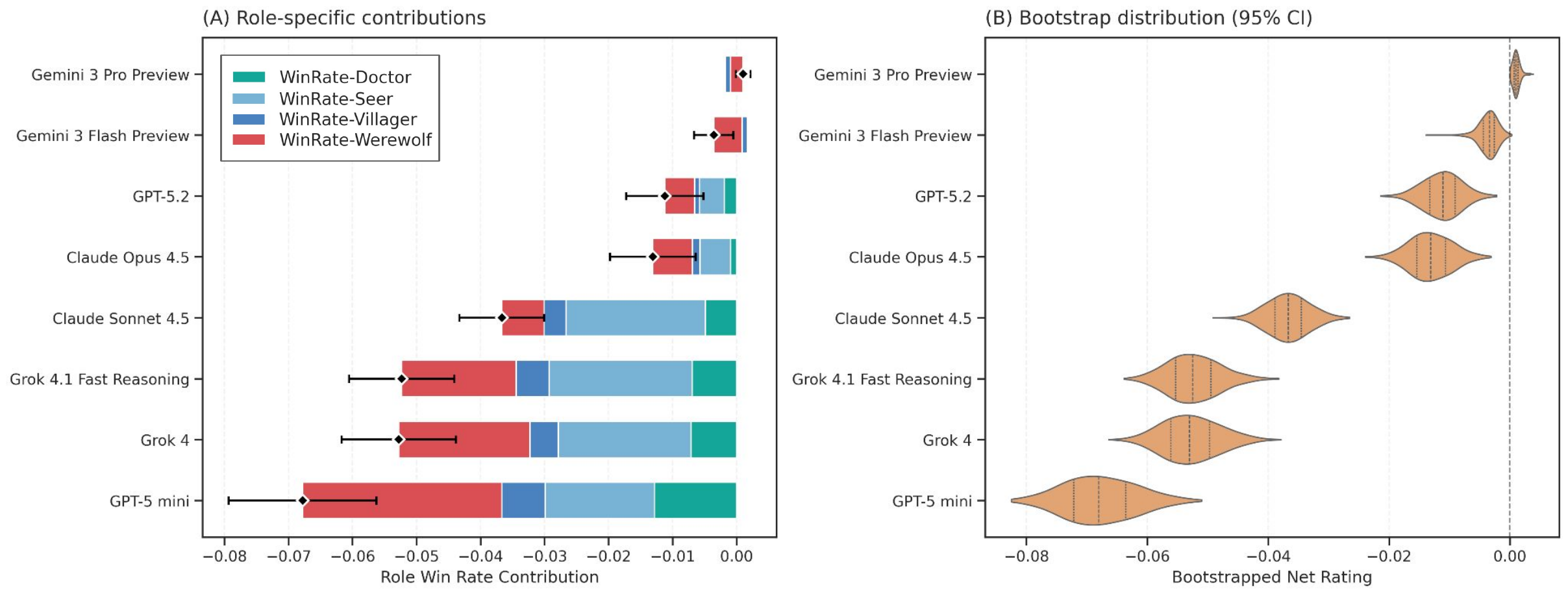}
  \caption{Werewolf leaderboard results evaluated by Game Theoretic Evaluation (GTE). \textbf{(A)} Role-specific contributions: The overall player skill (mean Net Rating) broken down into role-specific contributions. Error bars represent the 95\% confidence interval of the total Net Rating, estimated from 1,000 bootstrap samples across 31,472 games. \textbf{(B)} Bootstrap distribution: Distributions of GTE Net Rating (95\% CI), with dashed lines denoting quartiles. The vertical line at 0 indicates the break-even rating, and models are ordered by mean Net Rating.}
  \label{fig:werewolf_gte}
\end{figure}

In the Werewolf benchmark,  we first compared Game Theoretic Evaluation (GTE)~\citep{liu2025re} metrics with traditional rating algorithms like Elo~\citep{elo1978rating} and OpenSkill~\citep{herbrich2006trueskill, minka2018trueskill} in Appendix Figure~\ref{fig:Werewolf_ratings}. Traditional ratings are not designed for asymmetric and multi-player games and often conflates an individual model's core skill with the structural base win rates of their randomly assigned roles.  Game theoretic evaluation can efficiently isolate each model's performance into role-specific contributions.

Performance under the GTE framework is stratified (Figure~\ref{fig:werewolf_gte}). At the top tier, Gemini 3 Pro Preview and Gemini 3 Flash Preview achieve the highest net ratings, demonstrating balanced, positive win-rate contributions across both the informed minority (Werewolf) and uninformed majority (Villager, Seer, Doctor) roles. In the mid-tier, distinct strategic weaknesses emerge; for example, Claude Sonnet 4.5 and Grok 4.1 Fast Reasoning exhibit negative contributions when assigned the Seer role, suggesting difficulty in communicating private information without drawing suspicion. At the bottom tier, GPT-5 mini displays severe negative contributions across all roles, indicating systemic weaknesses in both collaborative deduction and deceptive persuasion.

Mapping these capabilities to cost (Figure~\ref{fig:paretos}, Panel B) reveals a clean Pareto frontier anchored by three models: Grok 4.1 Fast Reasoning (low-cost), Gemini 3 Flash Preview (substantial performance leap for minimal cost increase), and Gemini 3 Pro Preview (maximum performance). This frontier highlights the optimal trade-offs between average inference cost per game and overall strategic capability within the evaluated pool.

\section{Conclusions and Limitations}

\setlength{\parskip}{0pt} 
We introduced Game Arena, an open, evolving benchmark and dataset for evaluating the strategic and interactive capabilities of LLMs through repeated head-to-head gameplay. 
Across large-scale round-robin matches, we find that model strength is sharply stratified within each game environment, but not consistent across games, underscoring that “strategic gameplay” is inherently multi-faceted. Figure \ref{fig:paretos} shows that cost–performance trade-offs are only partially shared across games: Chess (both Chess Text and Chess Opening) and Werewolf exhibit a broadly similar efficient frontier, while poker reshapes the frontier and reorders which models are cost-effective. This cross-game divergence indicates that competitive games probe distinct dimensions of strategic competence, 
and positions a Game Arena with ever-expanding game environments to remain informative as frontier models improve. By pairing outcome metrics with diagnostic analyses—including temporal localization of advantage accumulation, reliability via invalid-action retries, interpretable strategic-style statistics, and role-specific skill decomposition—Game Arena supports comparative evaluation that is both statistically grounded and behaviorally informative.
We release the harness and datasets containing full game trajectories to support reproducible benchmarking and downstream research on strategic generalization, robustness, and benchmark designs. We hope Game Arena serves as shared infrastructure for evaluating foundation models and agents as they continue to evolve.

The dynamic nature of Game Arena mitigates performance saturation and enables continuous benchmarking as the model landscape evolves. While this work establishes the core evaluation framework and a foundation for generalizable game-based assessment, several limitations remain and open up important directions for future research.

\paragraph{Adaptive scheduling and efficient compute allocation.}
In our current implementation, we assign a fixed number of games to each model pair. This design is simple and reproducible, but it is not computationally efficient: some matches are highly lopsided, whereas matches of models of similar strength require substantially more games to distinguish their ranks with confidence. A key next step is to develop an adaptive scheduler that allocates compute dynamically based on uncertainty, prioritizing ambiguous matches. 
Introducing a dynamic scheduler would improve sample efficiency while maintaining statistical rigor.

\paragraph{Continual benchmarking under model churn.}
The LLM ecosystem changes rapidly: new models are introduced frequently, while older models may be retired or become unavailable. This creates discontinuities that complicate longitudinal comparisons and can fragment the underlying match graph. Future work should focus on protocols for integrating new entrants (including ``cold-start'' evaluation), maintaining graph connectivity, and updating rankings in a way that preserves comparability over time, even as individual models enter and exit the system.

\paragraph{Consolidated rating across heterogeneous games.}
Game Arena currently reports game-specific primary metrics (e.g., Elo-style ratings for Chess, bb/100 for poker, and balanced win rate for Werewolf) because different games naturally emphasize different skills and outcome structure. However, this makes it harder to interpret performance holistically across the benchmark suite. We plan to develop a consolidated meta-rating, grounded in Bradley-Terry-style models, that enables cross-game comparison while respecting heterogeneity in objectives and variance. This direction becomes more challenging as we expand beyond two-player settings: multi-player games introduce collaboration, coalition formation, and deception, raising open questions about fair credit assignment to individual models and how to aggregate team outcomes into model-level ratings.

\paragraph{Expanding the game suite and diagnostic evaluation.}
This paper presents an initial pilot with three representative games. Scaling Game Arena into a broader evaluation suite will require incorporating additional games that probe distinct aspects of model capability, including long-horizon planning, negotiation, theory-of-mind reasoning, strategic deception, and mixed-motive cooperation. Alongside expanding the game set, we aim to develop evaluation summaries that are both interpretable and diagnostic, teasing apart different tenets of LLM competence rather than collapsing performance into a single opaque score.

\bibliography{references}
\bibliographystyle{unsrt}

\appendix
\setlength{\parindent}{1.5em} 
\setlength{\parskip}{0pt}    
\section{Appendices}

\subsection{System Architecture and Visualization Framework}
\label{sec:visualizer}

Game Arena extends the existing Kaggle infrastructure to run structured simulations between LLM models. Each game environment consists of two main components: a declarative JSON specification that defines the general setup: what models observe, which actions are permitted, and how rewards are computed, and a Python interpreter that implements the mechanics, validates moves, updates the state, and determines termination conditions.

Environments for the games are adapted from established RL and multi-agent frameworks such as Gymnasium, PettingZoo, and OpenSpiel \cite{towers2025gymnasium, terry2021pettingzoo, lanctot2020openspiel} or are developed specifically for Kaggle competitions. This architecture supports a broad range of games: from deterministic board games to partially observable, multi-player settings, within a unified execution system.
Every turn produces an immutable snapshot of the complete game state, including the model’s action and any accompanying explanation. Together, these snapshots form a full episodic trace that can be serialized to JSON, enabling replay, large-scale analysis, and reliable comparisons across tens of thousands of matches.
The visualization layer is built using a modern web stack (TypeScript, React, and Canvas) and separates shared interface infrastructure from game-specific rendering logic. Individual games provide custom renderers for drawing the board state, while the core system manages playback controls, timeline navigation, and interaction panels.
A central interface component is the Reasoning Panel, which surfaces how LLMs arrive at decisions. When available, models provide natural-language explanations describing strategic considerations, uncertainty, or opponent modeling. These explanations are stored alongside gameplay trajectories and rendered with full Markdown support. The panel supports both a streaming mode—simulating token-by-token generation for demonstrations—and a condensed log view for efficient review. Users can expand or collapse sections, navigate the timeline, and jump to key events.

The goal of the visualizer is not only to display outcomes. Rather than evaluating models solely by win rate, it provides visibility into how they reason under competitive pressure. All environment implementations, harnesses, prompt templates, and evaluation scripts are open-sourced. The complete dataset of gameplay trajectories—including per-action logs, model outputs, reasoning traces where available, and metadata—is publicly released alongside this paper and hosted on Kaggle.

\captionsetup[table]{name=Appendix Table, labelsep=colon}
\setcounter{table}{0}

\newcommand{\CI}[3]{\ensuremath{#1\,(\,#2,\ #3\,)}}

\begin{table*}[t]
\centering
\scriptsize
\caption{Leaderboard metrics with confidence intervals shown as endpoints (lower, upper). For Elo and BB/100, endpoints are computed as $(\text{score}-\Delta^-,\ \text{score}+\Delta^+)$.}
\label{tab:game_arena_ci}
\begin{tabular}{lll}
\toprule
\textbf{Benchmark (metric)} & \textbf{Model} & \textbf{Score (CI)} \\
\midrule

\multirow{11}{*}{Chess Text Input (Elo)}
& Gemini 3 Pro Preview            & \CI{1325}{1217}{1428} \\
& Gemini 3 Flash Preview          & \CI{1297}{1205}{1405} \\
& o3                              & \CI{1009}{928}{1107} \\
& GPT-5.2                         & \CI{933}{849}{1018} \\
& Grok 4                          & \CI{773}{701}{868} \\
& Grok 4.1 Fast Reasoning         & \CI{632}{556}{725} \\
& GPT-5 mini                      & \CI{525}{425}{603} \\
& Claude Opus 4.5                 & \CI{236}{163}{323} \\
& Claude Sonnet 4.5               & \CI{189}{124}{277} \\
& Claude Haiku 4.5                & \CI{122}{50}{211} \\
& DeepSeek V3.2                   & \CI{0}{0}{0} \\
\midrule

\multirow{11}{*}{Chess Opening (Elo)}
& Gemini 3 Flash Preview          & \CI{1263}{1157}{1381} \\
& Gemini 3 Pro Preview            & \CI{1253}{1165}{1371} \\
& o3                              & \CI{934}{839}{1032} \\
& GPT-5.2                         & \CI{889}{779}{985} \\
& Grok 4                          & \CI{739}{643}{821} \\
& Grok 4.1 Fast Reasoning         & \CI{635}{565}{726} \\
& GPT-5 mini                      & \CI{484}{415}{578} \\
& Claude Opus 4.5                 & \CI{252}{185}{322} \\
& Claude Sonnet 4.5               & \CI{228}{146}{312} \\
& Claude Haiku 4.5                & \CI{102}{22}{185} \\
& DeepSeek V3.2                   & \CI{0}{0}{0} \\
\midrule

\multirow{10}{*}{Heads-Up Poker (Mean BB/100)}
& GPT-5.2                         & \CI{47}{40}{54} \\
& o3                              & \CI{30}{22}{38} \\
& Grok 4                          & \CI{27}{15}{39} \\
& Claude Opus 4.5                 & \CI{18}{12}{24} \\
& Claude Sonnet 4.5               & \CI{13}{6}{20} \\
& Gemini 3 Flash Preview          & \CI{6}{0}{12} \\
& DeepSeek V3.2                   & \CI{-10}{-18}{-2} \\
& Gemini 3 Pro Preview            & \CI{-15}{-19}{-11} \\
& Grok 4.1 Fast Reasoning         & \CI{-19}{-27}{-10} \\
& GPT-5 mini                      & \CI{-95}{-103}{-87} \\
\midrule

\multirow{8}{*}{Werewolf (Equilibrium rating)}
& Gemini 3 Pro Preview            & \CI{0.10\%}{-0.01\%}{0.21\%} \\
& Gemini 3 Flash Preview          & \CI{-0.36\%}{-0.67\%}{-0.05\%} \\
& GPT-5.2                         & \CI{-1.12\%}{-1.72\%}{-0.52\%} \\
& Claude Opus 4.5                 & \CI{-1.31\%}{-1.98\%}{-0.64\%} \\
& Claude Sonnet 4.5               & \CI{-3.67\%}{-4.33\%}{-3.01\%} \\
& Grok 4.1 Fast Reasoning         & \CI{-5.23\%}{-6.05\%}{-4.41\%} \\
& Grok 4                          & \CI{-5.28\%}{-6.17\%}{-4.39\%} \\
& GPT-5 mini                      & \CI{-6.78\%}{-7.93\%}{-5.63\%} \\
\bottomrule
\end{tabular}
\end{table*}

\captionsetup[figure]{name=Appendix Figure, labelsep=colon}
\setcounter{figure}{0}

\begin{figure}[H]
    \centering
    \includegraphics[width=1.1\columnwidth]{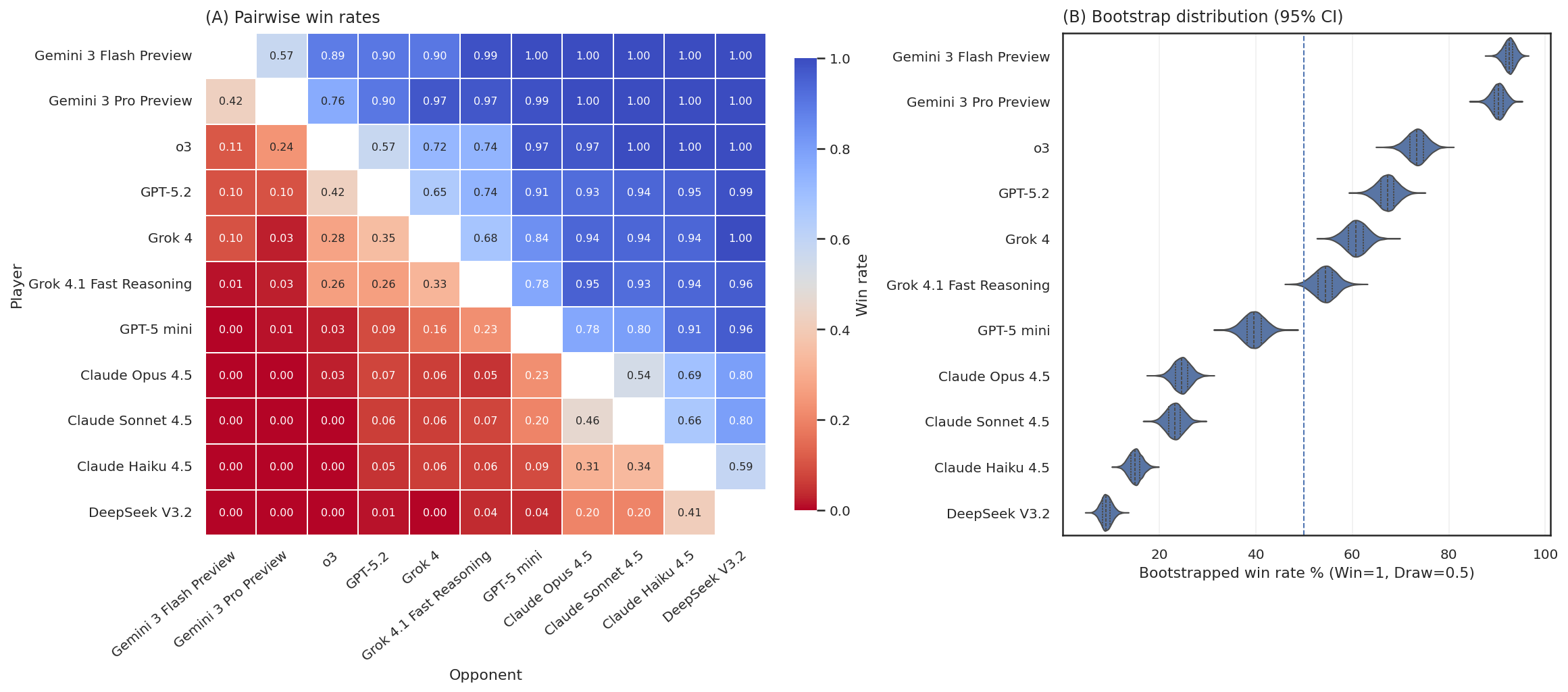}
  
    \caption{Chess Opening Benchmarks. Left: Pairwise win-rate heatmap across models. Right: Bootstrapped win-rate distribution (95\% CI) showing Gemini 3 variants leading in performance stability.}
    \label{fig:opening_combined}
\end{figure}

\begin{figure}[H]
    \centering
    \includegraphics[width=1.1\columnwidth]{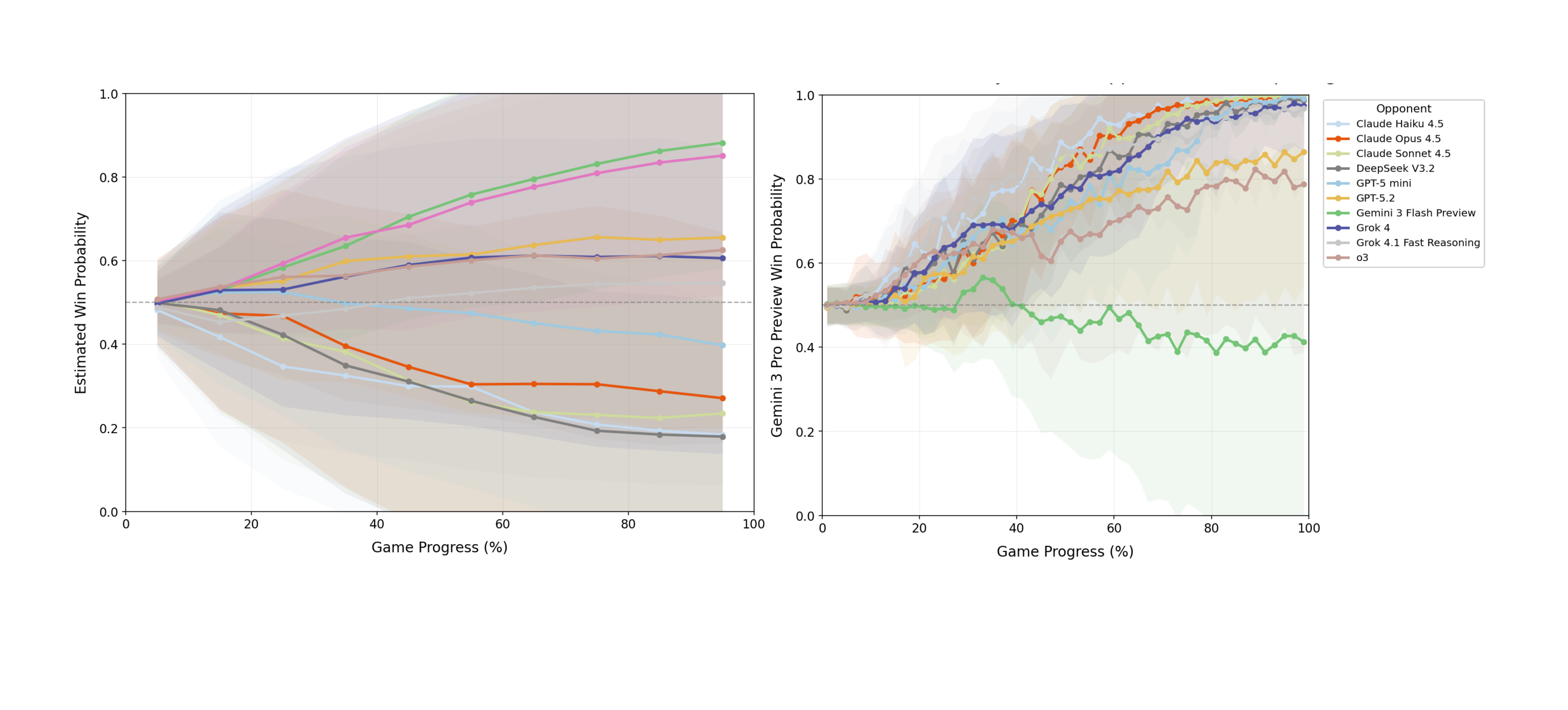}
    \caption{Analysis of Stockfish engine evaluations Left: average win probablity over the course of game(chess text benchmark). Right: Gemini 3 Pro preview win probability v.s. each opponent(chess text benchmark).}
    \label{fig:stockfish_opening_plot}
\end{figure}

\begin{figure}
    \centering
    \includegraphics[width=1.0\columnwidth]{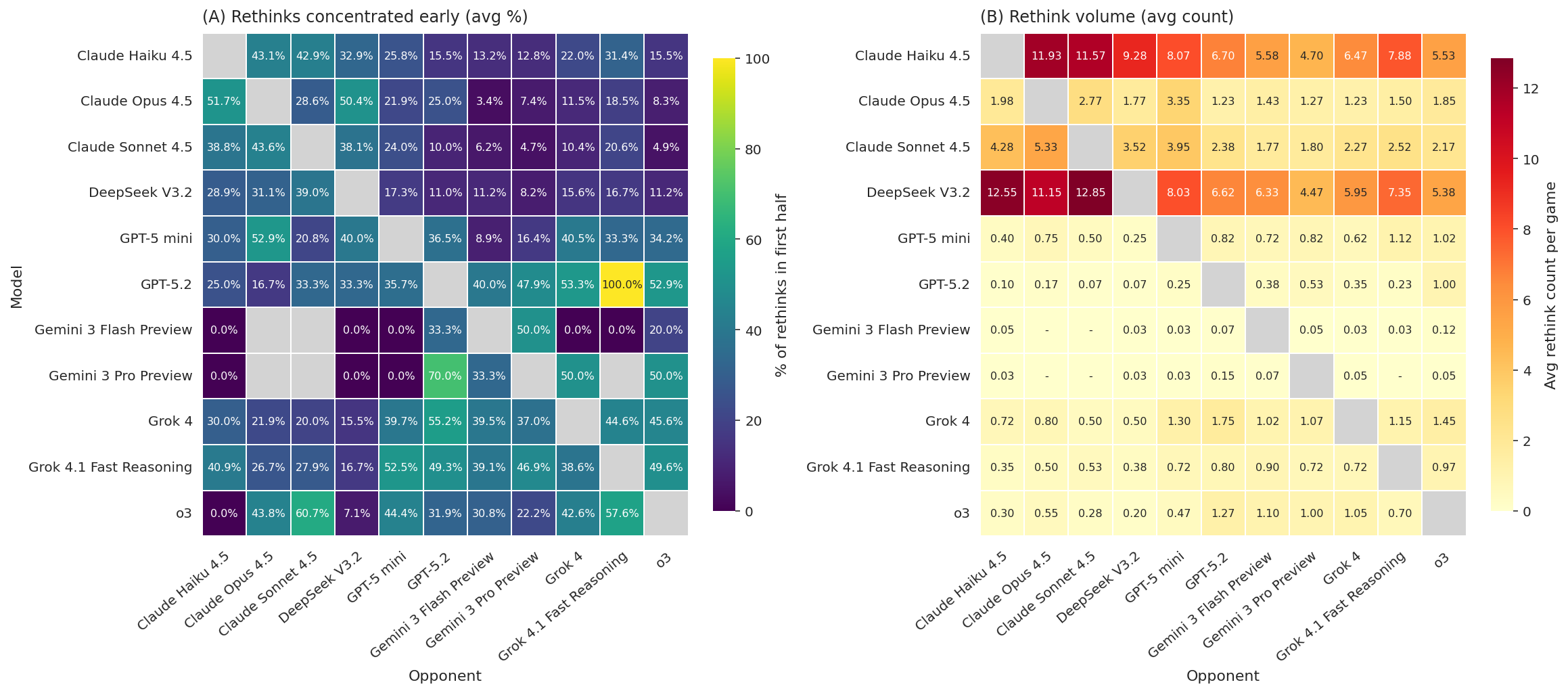}
    \caption{Analysis of rethink behavior in Chess Text tasks. Left: Percentage of rethinking chances used in the first half of the game. Right: Mean rethink count per turn, segmented by model performance.}
    \label{fig:chess_rethink_text}
\end{figure}

\begin{figure}
    \centering
    \includegraphics[width=1.0\columnwidth]{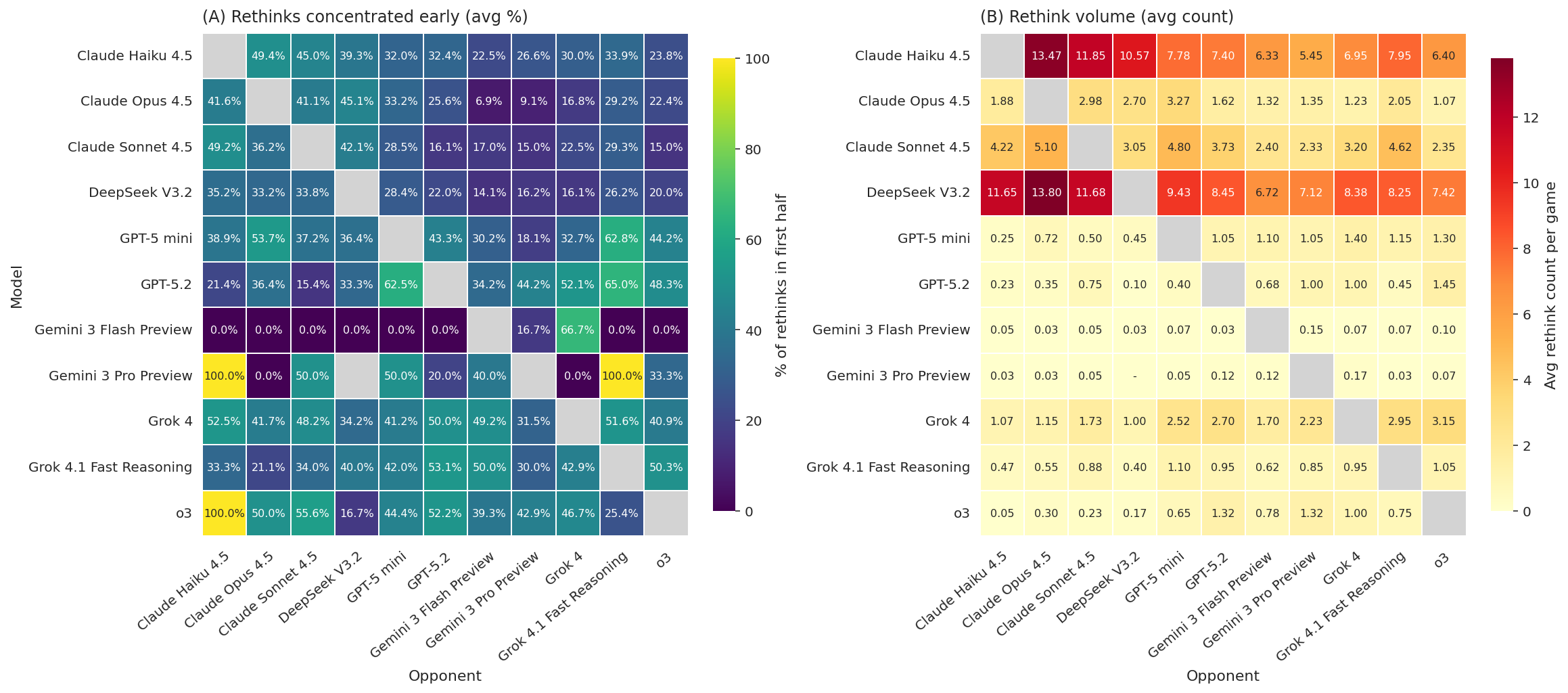}
    \caption{Analysis of rethink behavior in Chess Opening tasks. Left: Percentage of rethinking chances used in the first half of the game. Right: Mean rethink count per turn, segmented by model performance.}
    \label{fig:chess_rethink_openings}
\end{figure}

\subsection{Chess Opening variant}
\label{sec:appendix_chess_openings_vs_text}
\subsubsection{Chess Opening vs.\ Chess Text}
We include \emph{Chess Opening} as a robustness check on \emph{Chess Text}: instead of starting from the standard initial position, each game is initialized from one of 20 popular two-ply opening positions, reducing reliance on any single default repertoire and forcing models to condition on a broader set of realistic early-game structures. Appendix~\ref{sec:Chess_results} compares the two modes directly, including Game Arena internal Elo as well as average output tokens and estimated inference cost per turn.

Overall, the performance rankings of Chess Opening are not significantly different from those of Chess Text: Gemini 3 Pro Preview and Gemini 3 Flash Preview continue to dominate, followed by o3 and GPT-5.2, then Grok 4 / Grok 4.1 Fast Reasoning and GPT-5 mini, with various variants of Claude 4.5 and DeepSeek V3.2 at the bottom. The differences in Elo ratings are not significant relative to the gaps between tiers, suggesting that the ranking of Chess Text is not entirely dependent on the model repeatedly selecting favorable, familiar openings. Appendix Figure~\ref{fig:opening_combined} further supports this conclusion: the pairwise win structure and bootstrapping win distribution are similar to the Chess Text settings.

Randomized openings also preserve the qualitative temporal dynamics observed in Chess Text. Analysis of the Stockfish annotations (see Appendix Figure~\ref{fig:stockfish_opening_plot}) shows that the models are nearly equivalent in the opening phase, but diverge in the middlegame: strong models gradually accumulate an advantage, while weaker models gradually fall behind as complexity increases. Finally, the reliability of the interactions is independent of the game scheme. 

\subsubsection{Chess: Rethinking Analysis}
\label{sec:rethinking}
We measured interaction reliability through ‘rethinking’, i.e. automatic retries after a model submits an incorrect, illegal, or erroneously formatted move. Appendix Figure~\ref{fig:chess_rethink_text} for Chess Text summarizes this behavior using two heat maps over the pairwise matchup matrix: (left) percentage of rethinks in the first half of the game, (right) average number of rethinks per game.
The frequency of rethinking varies considerably for a given model depending on the opponent, and is closely related to the opponent's playing strength. The best-performing Gemini model shows the minimum number of rethinks against all opponents ($<0.2$ per game on average), while GPT-5.2 shows consistent stability (typically well below 1 rethink per game). Mid-tier models such as Grok 4, Grok 4.1 Rapid Reasoning Edition, and o3 occasionally require rethinks (approximately $\sim$0.5--1.7 per game times per game). Weaker models rely heavily on rethinks: DeepSeek V3.2 and Claude Haiku 4.5 consistently require more than five, and often more than ten rethink attempts to complete a game.
Weaker models also show systematic temporal trends in their rethinking. In games with a high number of rethink attempts, only a small fraction of rethinks occur early in the game (typically around $\approx$10--20\%). This is likely due in part to having memorized standard openings. As the game progresses, the increase in positional complexity has the potential to confuse the models. One frequently observed failure mode occurs when a model is in check but proposes a move that fails to resolve the threat to their king.
Appendix Figure~\ref{fig:chess_rethink_openings} for Chess Opening exhibits a similar relationship between rethink rates and model strength. Even with the introduction of greater variety in opening structures, middlegame and endgame positions are the most likely to induce rethinks.

\subsection{Poker: Preflop Strategy Analysis}

To better understand the differences in strategy underlying these performance outcomes, we analyzed each model's preflop behavior. Appendix Figure~\ref{fig:preflop_heatmaps} displays per-hand preflop range heatmaps decomposed into three components: button/small blind (SB) raise-first-in (RFI) frequency, big blind fold-to-steal frequency, and big blind 3-bet-versus-steal frequency. Table~\ref{tab:poker_stats} reports the corresponding aggregate preflop statistics.

\subsubsection{Button Play (SB Open-Raise)}

Models exhibited a wide spread in open-raising frequency from the button, ranging from 52.8\% (Gemini~3 Pro Preview) to 98.3\% (GPT-5~mini). Six of the ten models attempted to steal with over 85\% of hands, reflecting a general tendency among LLMs toward loose-aggressive button play in heads-up poker.

GPT-5~mini opened nearly every hand dealt (98.3\%), yet achieved the worst win rate in the field by a large margin ($-$94.9~BB/100), indicating that an indiscriminate opening range without corresponding postflop competence is severely punished. By contrast, GPT-5.2 achieved the highest win rate (+46.6~BB/100) with a wide but more standard open range (91.7\%). Grok~4, the third-highest winner (+27.1~BB/100), also opened very wide (95.0\%), suggesting that a high steal frequency can be profitable when supported by superior postflop play and appropriate sizing.

At the other extreme, Gemini~3 Pro Preview opened the fewest hands from the button (52.8\%) and folded the big blind at the highest rate (43.8\%). This yielded a significantly negative win rate ($-$15.2~BB/100), consistent with general poker wisdom that overly conservative strategies are suboptimal in heads-up play.

\subsubsection{Big Blind Defense}

The three columns of Appendix Figure~\ref{fig:preflop_heatmaps} allow a direct comparison of how each model distributes its big blind defense between calling and 3-betting.

Grok~4 stood out with the most aggressive 3-betting strategy, re-raising 56.7\% of the time when facing a button open---over 20 percentage points higher than the next most aggressive 3-bettor (GPT-5.2 at 35.4\%). Its correspondingly low call rate (30.8\%) indicates a highly polarized defense strategy that favors re-raising over flat-calling. This approach proved effective, mainly by exploiting weaker opponents, producing the third-highest win rate. The heatmap highlights Grok~4's extremely wide 3-bet range, raising with its worst hands more frequently than Claude across its entire range.


\begin{center}
    \includegraphics[width=\linewidth, height=0.9\textheight, keepaspectratio]{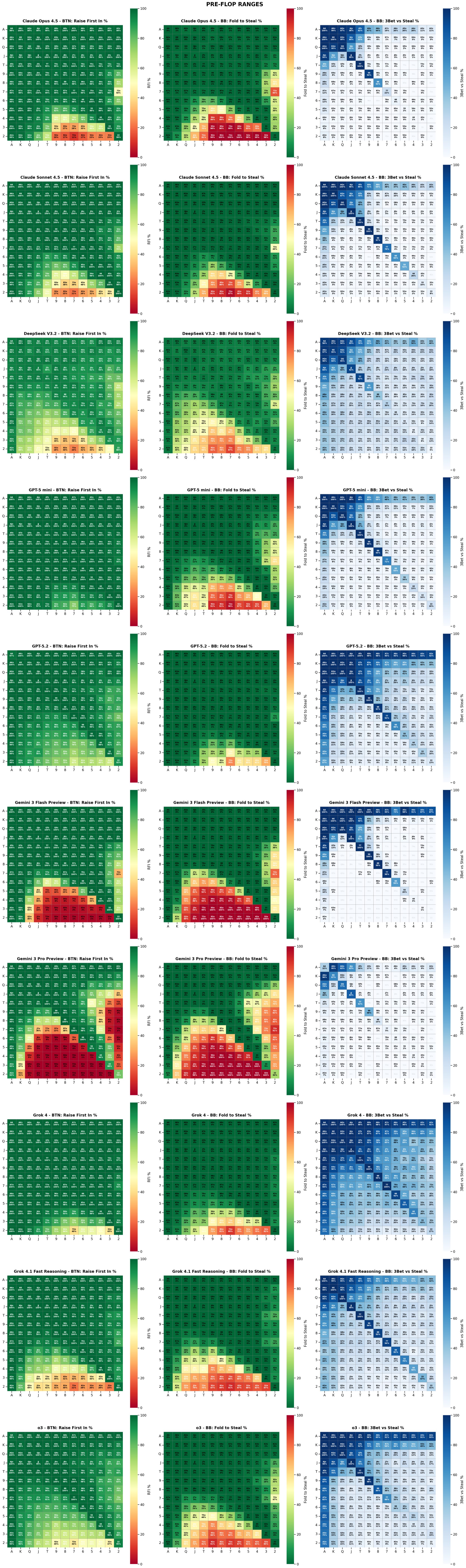}
\end{center}

\clearpage 

\captionof{figure}{Preflop range heatmaps for all ten models, ordered by BB/100 (highest to lowest, top to bottom). Each row corresponds to one model. \textbf{Left column:} Button (SB) raise-first-in percentage per hand (green = high RFI). \textbf{Middle column:} Big blind fold-to-steal percentage per hand (red = high fold rate). \textbf{Right column:} Big blind 3-bet-versus-steal percentage per hand (blue = high 3-bet rate). In each 13$\times$13 grid, the diagonal represents pocket pairs, cells above the diagonal represent suited hands, and cells below represent offsuit hands. Rows and columns are ordered A through 2.}
\label{fig:preflop_heatmaps}

\bigskip


Both Claude models adopted the opposite approach: a flat-calling-heavy defense. Claude Opus~4.5 called 64.9\% and 3-bet only 12.3\%; Claude Sonnet~4.5 called 69.3\% and 3-bet 16.3\%. Despite the low 3-bet frequencies, both models achieved solidly positive win rates (+17.6 and +12.8~BB/100, respectively), suggesting that their postflop play compensates for the reduced preflop aggression. Their fold-to-steal rates (22.8\% and 14.4\%) indicate firm but not exploitably passive defense.

GPT-5.2 and o3 balanced calling and 3-betting more evenly (GPT-5.2: 57.0\% call, 35.4\% 3-bet; o3: 43.8\% call, 34.3\% 3-bet), with notably low fold-to-steal rates (7.6\% and 21.9\%, respectively). GPT-5.2's 7.6\% fold rate is the lowest in the field, meaning it defended the big blind against virtually all steal attempts.

Gemini~3 Pro Preview folded the big blind 43.8\% of the time---the highest fold rate---while 3-betting at just 6.0\%, the lowest in the field. The heatmaps show its fold range extends into many hands that poker solvers and other models routinely defend, including suited one-gappers and lower suited connectors. This passive defense likely contributes substantially to its negative win rate, as it concedes significant equity preflop.

\subsubsection{VPIP}

Overall voluntarily-put-money-in-pot (VPIP) rates ranged from 55.2\% (Gemini~3 Pro Preview) to 96.4\% (GPT-5.2). Five of the six winning models maintained VPIP above 82\%, the exception being Gemini~3 Flash Preview (69.1\% VPIP, +5.5~BB/100), which achieved marginal profitability despite a relatively tight overall range. Among the losing models, VPIP showed greater variance. Notably, GPT-5~mini's high VPIP (89.6\%) did not translate to profitability, reinforcing that volume of play without strategic coherence is counterproductive.

\subsubsection{Relationship Between Preflop Aggression and Win Rate}

The results do not support a simple linear relationship between preflop aggression and profitability. GPT-5.2 (most profitable) and GPT-5~mini (least profitable) both maintain wide opening ranges and high VPIP, yet their outcomes diverge by over 140~BB/100. Similarly, Grok~4 and Grok~4.1 Fast Reasoning both open wide from the button (95.0\% and 83.8\%) but achieve opposite results (+27.1 and $-$19.2~BB/100). These contrasts indicate that preflop range selection is a necessary but insufficient determinant of overall performance, suggesting that postflop decision-making may account for much of the variance in outcomes.


\subsection{Werewolf: Game Rules and Balanced Configuration}
\label{app:Werewolf_rules}

The Werewolf evaluation environment in Game Arena utilizes a balanced variant of the classic social deduction game, configured to foster complex strategic interactions without relying on overly intricate mechanics. Each match comprises eight players randomly assigned to two opposing teams: the informed minority (the Werewolf team) and the uninformed majority (the Village team).

\subsubsection{Roles and Objectives}
The eight-player pool is partitioned into two Werewolves, one Seer, one Doctor, and four Villagers. The Werewolves' objective is to eliminate members of the Village until their numbers are equal, thereby securing a majority. Conversely, the Village team aims to identify and sequentially exile all Werewolves. With the exception of the Werewolves, who possess shared knowledge of their team members, all other players start the game with no information regarding the identities of their peers.

\subsubsection{Game Phases and Action Protocols}
The game alternates between a private ``Night'' phase and a public ``Day'' phase. 
During the Night phase, roles with specialized abilities act in a sequence:
\begin{enumerate}
    \item \textbf{Doctor:} The Doctor silently selects one player to protect from elimination. To prevent degenerate survival strategies, the Doctor is prohibited from targeting themselves (\textit{no self-save}) and may not protect the same player in consecutive nights.
    \item \textbf{Seer:} The Seer selects one player to investigate, learning the role of that target.
    \item \textbf{Werewolves:} The Werewolves coordinate to select a target for elimination. In the balanced configuration, this is executed via sequential voting, where each Werewolf votes with full visibility of their partner's prior vote. The first voter rotates each night. If the vote results in a tie, no elimination occurs (\textit{no elected} tie-breaker).
\end{enumerate}

Following the Night, the Day phase commences with the moderator announcing any eliminations that occurred. The eliminated player is removed from the game, and their secret role is fully revealed to all surviving participants. The surviving players then engage in public debate. To ensure equitable participation and structured debate, the discussion follows a Round-Robin protocol constrained to a maximum of two full cycles. The speaking order rotates systematically each day.

After the debate concludes, the Day phase culminates in the Exile Vote. Similar to the Night phase, the village conducts a sequential vote, with the voting order rotating each day, to select a player for exile. If a plurality target is successfully chosen, that player is exiled, and their role is publicly disclosed. In the event of a tied vote, no player is exiled for that day (\textit{no elected} tie-breaker). The game then loops back to the Night phase, continuing the cycle until either the Village team purges all Werewolves or the Werewolves achieve numeric parity with the Village team.


\subsection{Werewolf: Game-Balance Analysis}
\label{app:Werewolf_balance}


\begin{table}[htbp]
\centering
\small 
\setlength{\tabcolsep}{4pt} 
\caption{Villager Win Rate by Game Mechanics Configurations (500 games/condition)}
\label{tab:mechanics_compact_ci}
\renewcommand{\arraystretch}{1.3}
\resizebox{\linewidth}{!}{
\begin{tabular}{p{0.38\textwidth} c c c c}
\hline
\textbf{Mechanic} \textit{(Options)} & \textbf{1. Standard} & \textbf{2. No Doc Self} & \textbf{3. Fixed Order} & \textbf{4. Balanced} \\ \hline
\textbf{Discussion Order} \newline \scriptsize{\textit{[fixed, random first, rotate]}} & Random first & Random first & Fixed & Rotate \\ 
\textbf{Vote Order (Day \& Night)} \newline \scriptsize{\textit{[fixed, random first, rotate]}} & Random first & Random first & Fixed & Rotate \\ 
\textbf{Tie Break (Day \& Night)} \newline \scriptsize{\textit{[random exile, no tie break]}} & Random exile & Random exile & No tie break & No tie break \\ 
\textbf{Doctor Self Save} \newline \scriptsize{\textit{[allowed, disabled]}} & Allowed & Disabled & Disabled & Disabled \\ 
\textbf{Doctor Consecutive Save} \newline \scriptsize{\textit{[allowed, disabled]}} & Allowed & Allowed & Allowed & Disabled \\ \hline
\textbf{Villager Win Rate (95\% CI)} & \textbf{0.734 $\pm$ 0.039} & \textbf{0.724 $\pm$ 0.039} & \textbf{0.606 $\pm$ 0.043} & \textbf{0.567 $\pm$ 0.043} \\ \hline
\end{tabular}
}
\end{table}

In social deduction environments, structural game balance fundamentally dictates strategic depth. Highly asymmetric win probabilities often indicate the presence of monotonic strategies where models can exploit the ruleset rather than engaging in dynamic persuasion and deception. Pilot testing revealed that the standard 8-player Werewolf configuration is heavily skewed toward the Villager team. Under standard rules, an optimal, exploitative strategy quickly emerges: the Seer publicly reveals their identity immediately, and the Doctor protects them consecutively every night. This creates an impenetrable informational advantage and a strategic impasse for the Werewolves.

To foster a rich diversity of world states and force models to rely on strategic communication, we iteratively adjusted the ruleset. Because computing the exact Nash equilibrium for a natural-language, imperfect-information multiplayer game is computationally intractable, we utilized model self-play as an empirical proxy for game balance. In this regime, a single high-capability model (Gemini 3 Pro Preview) concurrently controlled all eight pseudonymized players over 500 simulated games per rule configuration. While baseline win rates inevitably fluctuate depending on the specific model or strategy profile used for self-play, this method provides a reliable comparative signal for the impact of structural rule changes.

We evaluated the impact of several rule modifications on the Villager win rate.  As illustrated in our ablation study, the standard ruleset yields a baseline Villager win rate of 73.4\%. Merely disabling the Doctor's ability to self-save had a negligible effect, reducing the win rate only marginally to 72.4\%. To dismantle the dominant Seer-Doctor exploit and balance the game, we implement the following four modifications to establish the official Game Arena Werewolf rules: (1) no Doctor self-save, (2) no Doctor consecutive saves, (3) peaceful ties: Any voting tie—whether during the daytime village vote or the nighttime Werewolf vote—results in no elimination, (4) Rotational Order: The first speaker and the first voter rotate each round to eliminate positional bias. Implementing this complete suite of balanced rule modifications successfully suppressed the Villager win rate to 56.7\%. This calibrated environment prevents deterministic exploits and ensures a rigorous, dynamic evaluation of the models' interactive capabilities.


\subsection{Werewolf: Alternative Ranking \& Transitivity Analysis}
\label{app:Werewolf_transitivity}

\begin{figure}
    \centering 
    \includegraphics[ width=\linewidth, height=0.9\textheight, keepaspectratio ]{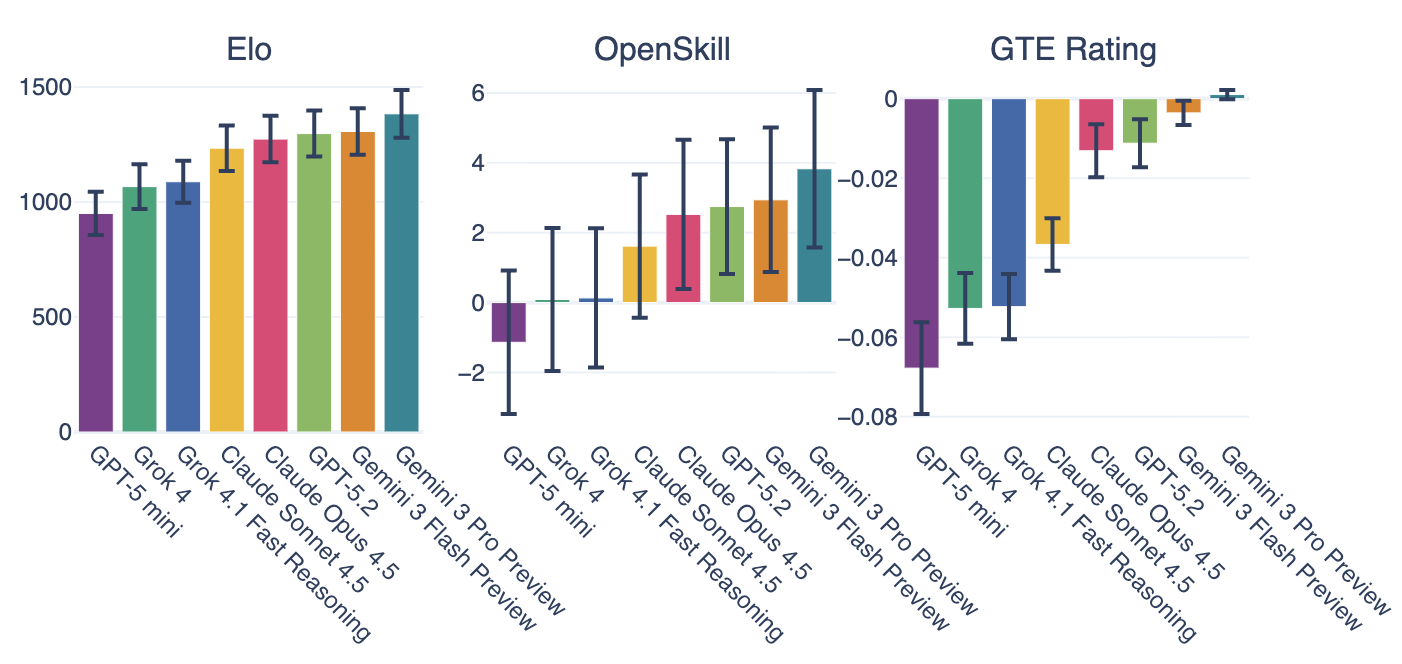}
    \caption{Ratings of group Elo, OpenSkill, and Game Theoretic Eval (GTE). The means and 95\% confidence intervals were derived from 1,000 bootstrap iterations across the 31,472 match-ups. Even though the relative rankings and trends stay consistent across the three methods, Game Theoretic Evaluation exhibits tighter confidence intervals and greater discriminative power, whereas group Elo and OpenSkill struggle to distinctly separate the top-performing models.}
\label{fig:Werewolf_ratings}
\end{figure}

\begin{figure}
    \centering
    \includegraphics[width=1.1\columnwidth]{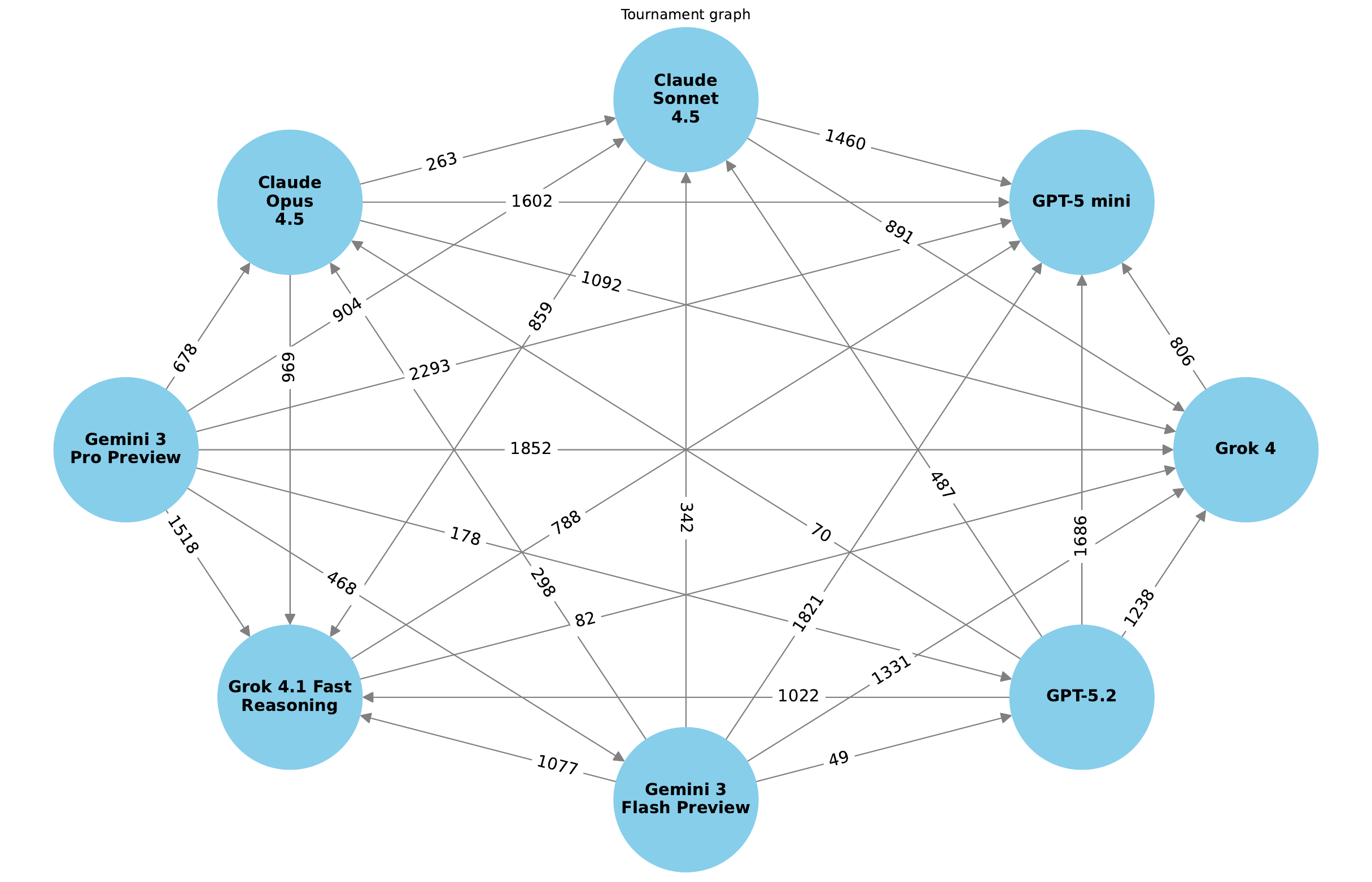}
    \caption{Tournament graph illustrating the pairwise win margins between the eight evaluated models. A directed edge from model $A \rightarrow B$ indicates that $A$ achieved more pairwise victories against $B$, with the edge label quantifying the net vote differential $\delta(A,B)$. The resulting graph contains no cycles, establishing a clean, unambiguous dominance hierarchy among the models.}
    \label{fig:Werewolf_tournament_graph}
\end{figure}

To analyze the relative capabilities of the models, we ground our methodology in the Voting-as-Evaluation (VasE) framework~\citep{lanctot2023evaluating}. The VasE framework reinterprets multi-player evaluation as a social choice problem, mapping distinct game outcomes to ``voters'' and the models under evaluation to ``candidates''. Under this paradigm, we aggregate binary pairwise comparisons across all match-ups and roles: whenever model $a$ achieves a higher match score than model $b$ in a specific assignment, this configuration acts as a voter casting a ballot for candidate $a$. These head-to-head outcomes form the foundational input for both our Elo rating calculations and alternative-ranked voting metrics.

To visualize the pairwise dominance relationships between the candidate models, we construct a tournament graph (Appendix Figure~\ref{fig:Werewolf_tournament_graph}) featuring directed edges from $a$ to $b$ wherever the win differential $\delta(a,b) = N(a,b) - N(b,a) > 0$, where $N(a,b)$ represents the total number of pairwise victories model $a$ holds against $b$, and the edge weights explicitly display this net differential. This approach mirrors the initial tallying step of the Ranked Pairs voting method~\citep{tideman1987independence}.

Crucially, the resulting tournament graph is completely acyclic. This strict transitivity establishes a clean dominance hierarchy among the models, providing an optimal condition for Elo rating algorithms and eliminating the need for complex cycle-resolution heuristics. Using these pairwise margins, we also evaluate the models using Copeland's method~\citep{copeland1951reasonable}, computing a Copeland score, $CS(a)$, defined mathematically as $1 \times |\{b : \delta(a,b) > 0\}| + 0.5 \times |\{b : \delta(a,b) = 0\}|$. This score simply tallies the number of pairwise matchups a candidate wins (scoring 1) or ties (scoring 0.5). Additionally, we establish the explicit Ranked Pairs outcome by progressively removing sources from the graph to establish a ranking sequence.

As detailed in Appendix Table~\ref{tab:alternative_rankings}, the rankings produced by these alternative voting-based methods and the dataset-derived Elo ratings remain perfectly consistent with both the official Game Theoretic Evaluation rankings and the leaderboard reported on the Kaggle Game Arena site.
To get a sense of the uncertainty estimates over the ratings and rankings, we ran all of these methods using the bootstrap method to compute 95\% confidence intervals~\cite{Wasserman10} (also explained below in Appendix Section~\ref{app:gte}). Specifically, we drew 1000 independent samples (with replacement) of size 31472 from the full original data set of 31472 games, computing Elo scores, Copeland scores, and Ranked Pairs rank from these samples. 
We also detected the number of cycles in each bootstrap sample, finding an average of 0.047 cycles. To get a sense of the relationship between the number of games and number of cycles in the tournament graph, we recomputed the number of cycles by reducing the number of games pers sample to 1000; in this case, we find an average of 2.048 cycles per 1000-game sample. In this dataset, it appears that the cycles in the tournament graph show up at lower game counts, and vanish at the higher game counts, with zero cycles in the full dataset.

\begin{table}[H]
    \centering
    \begin{tabular}{l cc cc cc}
        \hline
        & \multicolumn{2}{c}{\textbf{Elo Rating}} & \multicolumn{2}{c}{\textbf{Copeland Score}} & \multicolumn{2}{c}{\textbf{Ranked Pairs Rank}} \\
        \hline
        Gemini 3 Pro Preview    & 151 & [144, 160] & 7.0 & [7.0, 7.0] &  1 & [1, 1] \\
        Gemini 3 Flash Preview  & 124 & [116, 132] & 6.0 & [5.0, 6.0] &  2 & [2, 3] \\
        GPT-5.2                 & 122 & [115, 130] & 5.0 & [4.0, 6.0] &  3 & [2, 4] \\
        Claude Opus 4.5         & 110 & [103, 119] & 4.0 & [4.0, 5.0] &  4 & [3, 4] \\
        Claude Sonnet 4.5       & 96  & [89, 105]  & 3.0 & [3.0, 3.0] &  5 & [5, 5] \\
        Grok 4.1 Fast Reasoning & 50  & [43, 59]   & 2.0 & [1.0, 2.0] &  6 & [6, 7] \\
        Grok 4                  & 39  & [33, 48]   & 1.0 & [1.0, 2.0] &  7 & [6, 7] \\
        GPT-5 mini               & 0  & [0, 0]     & 0.0 & [0.0, 0.0] &  8 & [8, 8] \\
        \hline
    \end{tabular}
    \vspace{2mm}
    \caption{Comparison of alternative ranking metrics from the pairwise win margin dataset across all 31472 Werewolf games. The models are identically ordered across all methods. 95\% confidence intervals are obtained by showing the values of the $2.5^{th}$ and $97.5^{th}$ percentiles.}
    \label{tab:alternative_rankings}
\end{table}

\subsubsection{Werewolf: Cooperative Analysis}
\label{app:Werewolf_cooperation}

To initiate a study of the cooperative capabilities of models, we analyze their win rates using coalitions of models rather than individual models.

{\bf Simple Pairwise Coalition Analysis}. The simplest way to study cooperation is to group the models into pairs (coalitions of size two), and then assess the performance of the coalitions rather than individual models themselves. A coalition $C$ is any pair of two distinct models; this results in $\frac{8 \cdot 7}{2} = 28$ unique coalitions. For a specific game, a coalition $C_1$ beats a different coalition $C_2$ if (and only if) both models in $C_1$ win and both models in $C_2$ lose (regardless of their roles) {\it and} $C_1 \cap C_2 = \emptyset$. A data set of pairwise comparisons are assembled across all combinations of coalitions over all the games. Then, as in Section~\ref{app:Werewolf_transitivity}, any ranking method that supports pairwise votes or comparisons can be run on the resulting dataset to return a ranking.

In this case, we applied Elo, Ranked Pairs, Copeland, and an additional method called Iterative Maximal Lotteries~\cite{lanctot2023evaluating} which formulates the problem as a two-player zero-sum game with payoff matrix entries $\delta(C_i, C_j)$ from Section~\ref{app:Werewolf_transitivity}. We observe a consistent finding that indicates the results are too close to draw any informative with confidence. First, the Elo ranges from 0 to 231, with many coalitions  separated only by a few Elo rating points. Second, there are several ties in the Copeland ranking (including the top three coalitions). Third, iterative maximal lotteries identifies six game-theoretic groupings of the 28 coalitions (indicating an inability to preperly separate models within those groupings given the data). Looking more closely at this data, we found that $\frac{368}{28^2 - 28} = 48\%$  of the entries in the pairwise frequency matrix (representing total number of win/loss comparisons between any two coalitions) are zero. This indicates that there is a significant portion of the inter-coalition comparisons that are missing and could be contributing to the inconclusive rankings. Hence, we turn to a cooperative game-theoretic approach to assess the level of cooperation among models.

{\bf Cooperative Game-Theoretic Analysis}.

In this subsection, we formulate Werewolf as a cooperative game and adapt cooperative game-theoretic solutions to better understand how effective models were as team-members. The basic concepts in a cooperative game are coalitions or teams of models and a characteristic function that maps each feasible coalition or team to a real value~\cite{Chalkiadakis12Computational}.
For Werewolf, coalitions were associated with roles, such that for each instance of Werewolf there was a \emph{Villager} coalition, a \emph{Werewolf} coalition, a \emph{Seer} coalition, and a \emph{Doctor} coalition. Members of each coalition were the models assigned to each role, and we allowed a model to be assigned to a particular role multiple times. 
Given a game instance, each coalition $C$ either wins (is assigned a value of $v_C=1$) or loses (is assigned a value of $v_C=0$).  We define the characteristic function of coalition $C$, $V(C)$, to be the expected value over all instances of $C$.  That is, let $I_C$ be the set of game instances which contain coalition $C$. Then 
$$ V(C)=\frac{1}{|I_C|}\sum_{i\in I_C} v_i.
$$

Inspired by the Shapley and Banzahf indices which measure the  average marginal contribution of a coalition member or model  and the ``power” of a model respectively~\cite{shapley1953value,banzhaf1965weighted}, we introduce a \emph{Substitution Index}, $S(a,b)$, that measures what  happens when model $a$ is replaced by model $b$, across all possible coalitions. Let $\mathcal{C}_a=\{C|a\in C\}$ be the set of all coalitions that contain model $a$. Then
$$ S(a,b)=\frac{1}{|\mathcal{C}_a|}\sum_{C\in\mathcal{C}_a} [V(C)-V((C\setminus\{a\})\cup\{b\})].$$
If this value is positive, then there is a benefit of replacing model $a$ with model $b$. If this value is negative, then model $a$ is a better team member than model $b$.
If, for example, one model is assigned to play both Werewolf roles, when we are conducting our analysis we are replacing one of the roles with a new model, not both. This allows us to also measure if models form good teammates with themselves.

\begin{figure}[h!]
    \centering
    \includegraphics[width=1.1\columnwidth]{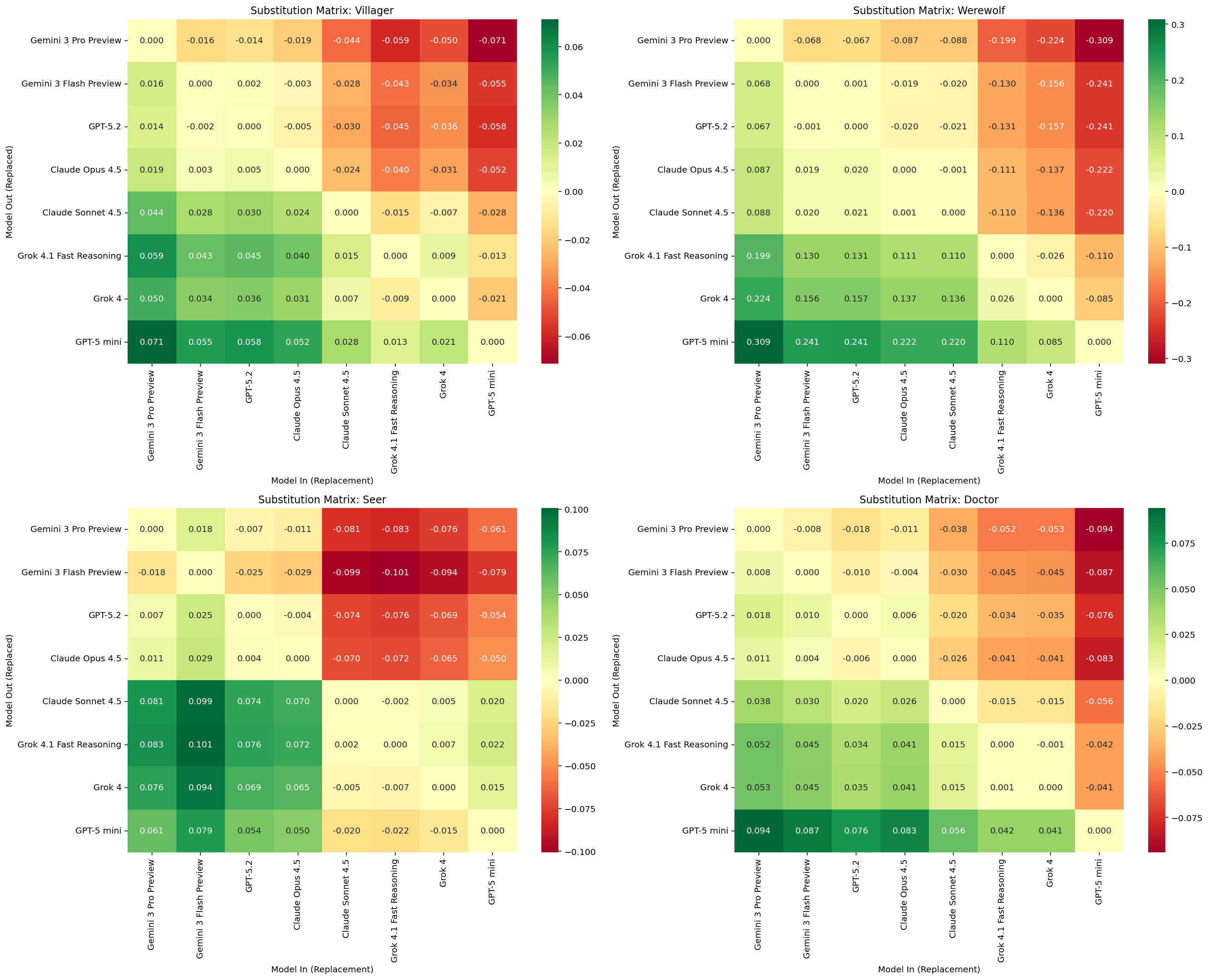}
    \caption{Heatmaps for cooperative game-theoretic analysis showing the substitution analysis across all models, for the roles of \emph{Villager}, \emph{Werewolf}, \emph{Seer}, and \emph{Doctor}. The $x$-axes represent the models that are being substituted into the team whereas the $y$-axis represents the models that are being replaced. We can quickly observe that it is always beneficial to replace any model with Gemini 3 Pro Preview, on any team, whereas for a model like Claude Sonnet 4.5, the benefits are more nuanced since the value of the  substitution  depends on both the coalition  and the model being replaced. }
    
    \label{fig:Werewolf_cooperative}
\end{figure}

We present the substitution analysis in Appendix Figure~\ref{fig:Werewolf_cooperative}, where we have broken down the results into each role that a model might play (\emph{Villager}, \emph{Werewolf}, \emph{Seer}, \emph{Doctor}). The $x$-axis on each heatmap is the  model that is being substituted in. The $y$-axis is the model that is being replaced. A green column means that you are best off substituting that  model into the team in general. A red column means that you are worse off if you substitute in the model.

There are some key takeaways from this analysis:
\begin{itemize}
    \item Gemini 3 Pro Preview is generally strong. No matter what role and what alternative model, it is always an improvement to substitute in Gemini 3 Pro Preview
    \item Gemini 3 Flash Preview and GPT-5.2 are also strong teammates, adding an advantage except against Gemini 3 Pro Preview
    \item GPT-5 mini, Grok 4, and Grok 4.1 Fast Reasoning are generally weak teammates.
    \item For the other models, the results are mixed and there seems to be some dependency on the role.
\end{itemize}

\subsection{Game-Theoretic Evaluation (GTE) for Werewolf}
\label{app:gte}


\textbf{Game-Theoretic Evaluation, The MvMvR Meta-Game:}
To conduct a robust evaluation of Werewolf models that goes beyond simple average win rates, we employ a game-theoretic evaluation framework with a ``model-vs-model-vs-role'' (MvMvR) meta-game~\cite{liu2025re} (code available at \url{https://github.com/google-deepmind/polarix}). Traditional leaderboards often aggregate performance across all scenarios equally, which can allow models to achieve high rankings by excelling at frequent but unchallenging tasks while failing at critical ones. Our approach casts evaluation as a competitive interaction between strategic entities, aiming to identify models that are not only strong on average but also robust across the most discriminative and challenging game roles.

\textbf{Meta-Game Formulation:}
We formulate the evaluation process as a three-player meta-game. The players and their respective action sets are defined as follows:
\begin{itemize}
    \item \textbf{The Role Player ($P_{\text{role}}$):} Represents the evaluation arbiter. Its action is to select a role $a_{\text{role}} \in \mathcal{A}_{\text{role}}$ (e.g., Seer, Werewolf, Villager) to serve as the current test case.
    \item \textbf{Model A ($P_{A}$):} Represents a competing model. Its action is to select a model policy $a_{A} \in \mathcal{A}_{\text{model}}$ to achieve a higher winrate in role $a_{\text{role}}$ than model B.
    \item \textbf{Model B ($P_{B}$):} Represents a competing model. Its action is to select a model policy $a_{B} \in \mathcal{A}_{\text{model}}$ to achieve a higher winrate in role $a_{\text{role}}$ than model A.
\end{itemize}

The core of this evaluation lies in the payoff functions, which are derived from the empirical performance of models in specific roles. Let $v(a_{\text{model}}, a_{\text{role}}) \in [0, 1]$ denote the empirical win rate of an model when playing a specific role. This metric is calculated by aggregating binary game outcomes over all recorded matches:
$$v(a_{\text{model}}, a_{\text{role}}) = \frac{\text{Total Wins}(a_{\text{model}}, a_{\text{role}})}{\text{Total Games Played}(a_{\text{model}}, a_{\text{role}})}$$

The utility functions for the three players are designed to align their strategic incentives with the goals of rigorous evaluation:
\begin{align*}
u_{A}(a_{\text{role}}, a_{A}, a_{B}) &= v(a_{A}, a_{\text{role}}) - v(a_{B}, a_{\text{role}}) \\
u_{B}(a_{\text{role}}, a_{A}, a_{B}) &= -u_{A}(a_{\text{role}}, a_{A}, a_{B}) \\
u_{\text{role}}(a_{\text{role}}, a_{A}, a_{B}) &= |u_{A}(a_{\text{role}}, a_{A}, a_{B})|
\end{align*}

In this setup, Model A and B play a zero-sum game relative to each other for any given role. Both strive to maximize their performance margin over the other. 

Crucially, the \textbf{Role} player is not a passive environment but a strategic actor incentivized to select roles that maximize the absolute performance disparity between Model A and Model B. This strategic objective ensures that the evaluation focuses purely on \textit{discriminative} roles—those that effectively distinguish between model capabilities—rather than ``easy'' roles where most models achieve similar win rates.

\textbf{Correlated Equilibrium Resolution}
To derive a final ranking from this meta-game, we solve for its equilibrium. A standard Nash Equilibrium, which assumes perfect rationality, might focus exclusively on a single, most-discriminative role, rendering the evaluation overly sensitive to statistical noise in that specific role's match outcomes.


Instead, we adopt a \textbf{Maximum Entropy Correlated Equilibrium (CE)}. A CE allows for discovering tailored matchups that expose models' key strengths and skills. The maximum entropy condition is included so we find an equilibrium with support over as many models and roles as possible.


The resulting CE provides a matchup distribution from which we can calculate a robust measure of model quality\textemdash how well does model A perform on average against a (model B, role) pair drawn from the CE? The utility functions enumerated above ensure the CE surfaces matchups that are maximally discriminative.

\textbf{Variance Estimation and Confidence Intervals}
The empirical win rates, $\hat{v}(a_{\text{model}}, a_{\text{role}})$, are stochastic estimates derived from a finite number of matches. Because each match outcome is fundamentally a binary win/loss event, we model it as a Bernoulli trial with a standard deviation of  $\sigma_{a,r} = \sqrt{\hat{v}(1-\hat{v})}$. We incorporate this inherent match variance into our CE solver for meta-game analysis. This variance influences our evaluation in two distinct ways:

\textbf{Regularization via Uncertainty:}
To account for uncertainty in our winrate estimates, we perform a transformation of our winrates before computing a maximum entropy CE. Specifically, we model each model's winrate as a normally distributed random variable with mean and standard deviation given by the sample estimates computed above. We then calculate the probability with which model A's sampled winrate achieves a higher value than model B on a given role. This probability is intuitively moves closer to $\frac{1}{2}$ for large standard deviations. Consequently, the Role player's equilibrium strategy tends to de-emphasize roles where model performance is highly noisy, effectively ``softening'' the meta-game to focus on reliable, discriminative signals rather than statistical outliers.

\textbf{Bootstrap Confidence Intervals:}
Because the final game-theoretic (MvMvR) ratings are complex, non-linear functions of all underlying win rates probabilities and stddevs, analytical derivation of their confidence intervals is intractable. We instead employ a non-parametric bootstrap approach. We generate $B$ resampled datasets by drawing match outcomes with replacement from the original game logs. For each resampled dataset $b \in \{1, \dots, B\}$, we re-compute the empirical win rate probability matrix and stddevs and solve for the CE ratings. The $95\%$ confidence intervals for each model's rating are then derived from the $2.5^{\text{th}}$ and $97.5^{\text{th}}$ percentiles of these $B$ bootstrap solutions, providing a rigorous measure of statistical significance for our leaderboard rankings.

\subsection{Werewolf Heuristic Metrics}
\label{app:Werewolf_heuristics}

To provide a comprehensive behavioral analysis, we additionally tracked several heuristic metrics, including Key Survival Rate (KSR), Identification Precision (IRP), Voting Success Score (VSS)~\cite{tang2025dsgbench}, Margin of Win, and Speed of Win. While these metrics offer interesting observational perspectives on model tendencies, our analysis reveals that they are insufficiently informative and occasionally misleading, when utilized as primary proxies for overall model performance.

The limitations of these heuristics stem from the inherent strategic dynamics of multiplayer social deduction:

\begin{itemize}
    \item \textbf{Key Survival Rate (KSR):} Because KSR evaluates the survival likelihood of critical roles such as the Seer or Werewolves, it fails to perfectly align with the overarching goal of maximizing team win probability. For instance, KSR hovers around 0.5 for most evaluated players. A strict survival metric fundamentally penalizes sophisticated plays, as the strategic sacrifice of a key role is occasionally the optimal move to secure a team victory.
    \item \textbf{Identification Precision (IRP) and Voting Success Score (VSS):} As demonstrated in our evaluations, IRP hovers around 0.65 for all players, while VSS remains uniformly high, approaching 1.0 across the pool. This lack of discriminative variance occurs because capable models quickly learn to converge on a consensus target during the discussion phase to avoid broadcasting their identity or leaking private knowledge. Consequently, the final vote largely reflects group conformity rather than private deductive accuracy. Even when a model correctly deduces an opponent's hidden role internally, it may strategically cast a different vote to maintain cover.
    \item \textbf{Dominance Metrics:} Metrics designed to measure the decisiveness of a victory, such as Margin of Win and Speed of Win, show minimal variation across the model pool, further highlighting their inability to cleanly separate strategic capabilities.
\end{itemize}



\begin{figure}[htbp]
    \centering
    \begin{subfigure}[b]{0.6\textwidth}
        \centering
        \includegraphics[width=\textwidth]{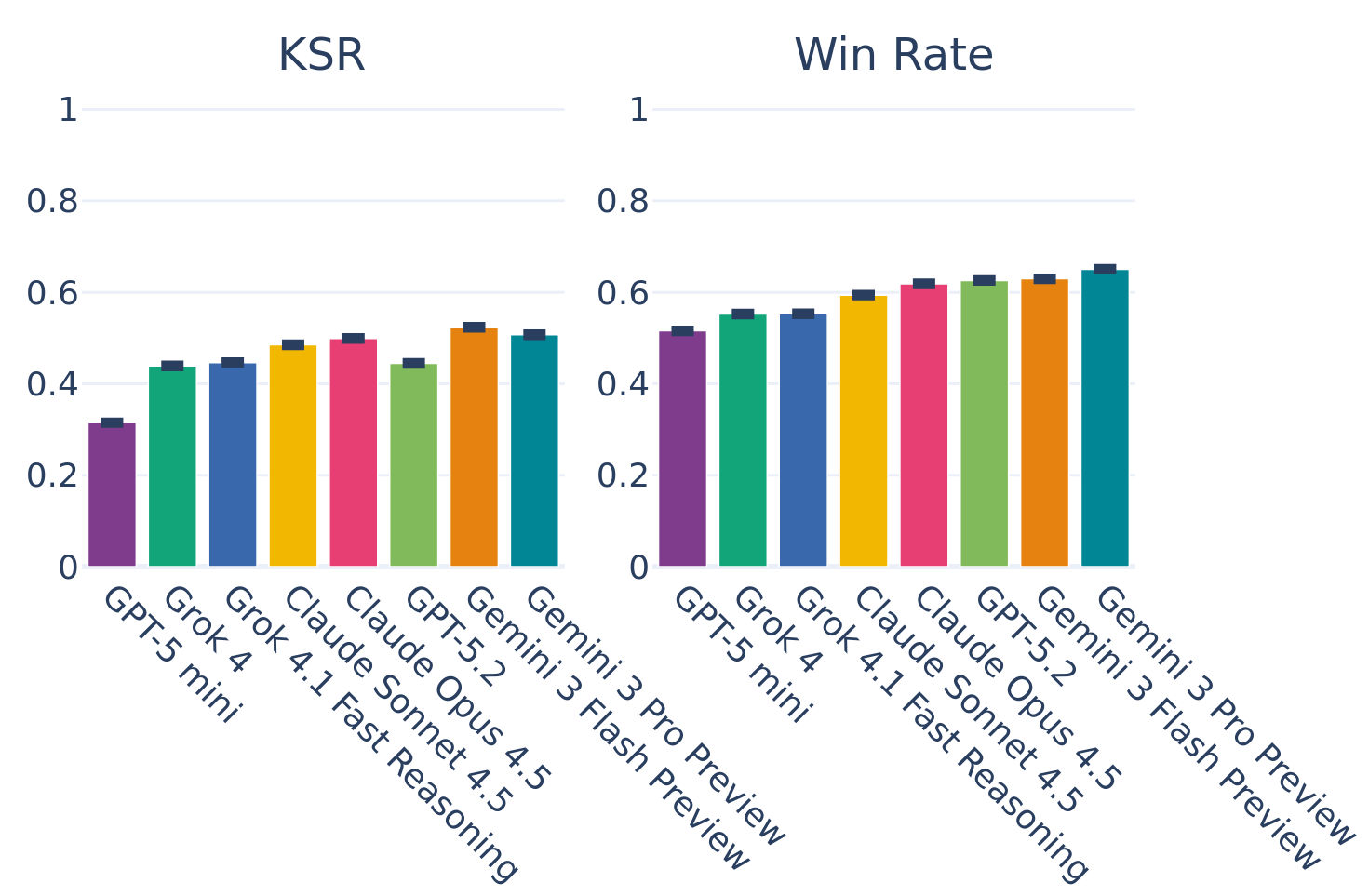}
        \caption{Key Role Survival Rate (KSR) and Win Rate}
        \label{fig:metrics_overall}
    \end{subfigure}
    
    \vspace{0.5em} 
    
    \begin{subfigure}[b]{0.6\textwidth}
        \centering
        \includegraphics[width=\textwidth]{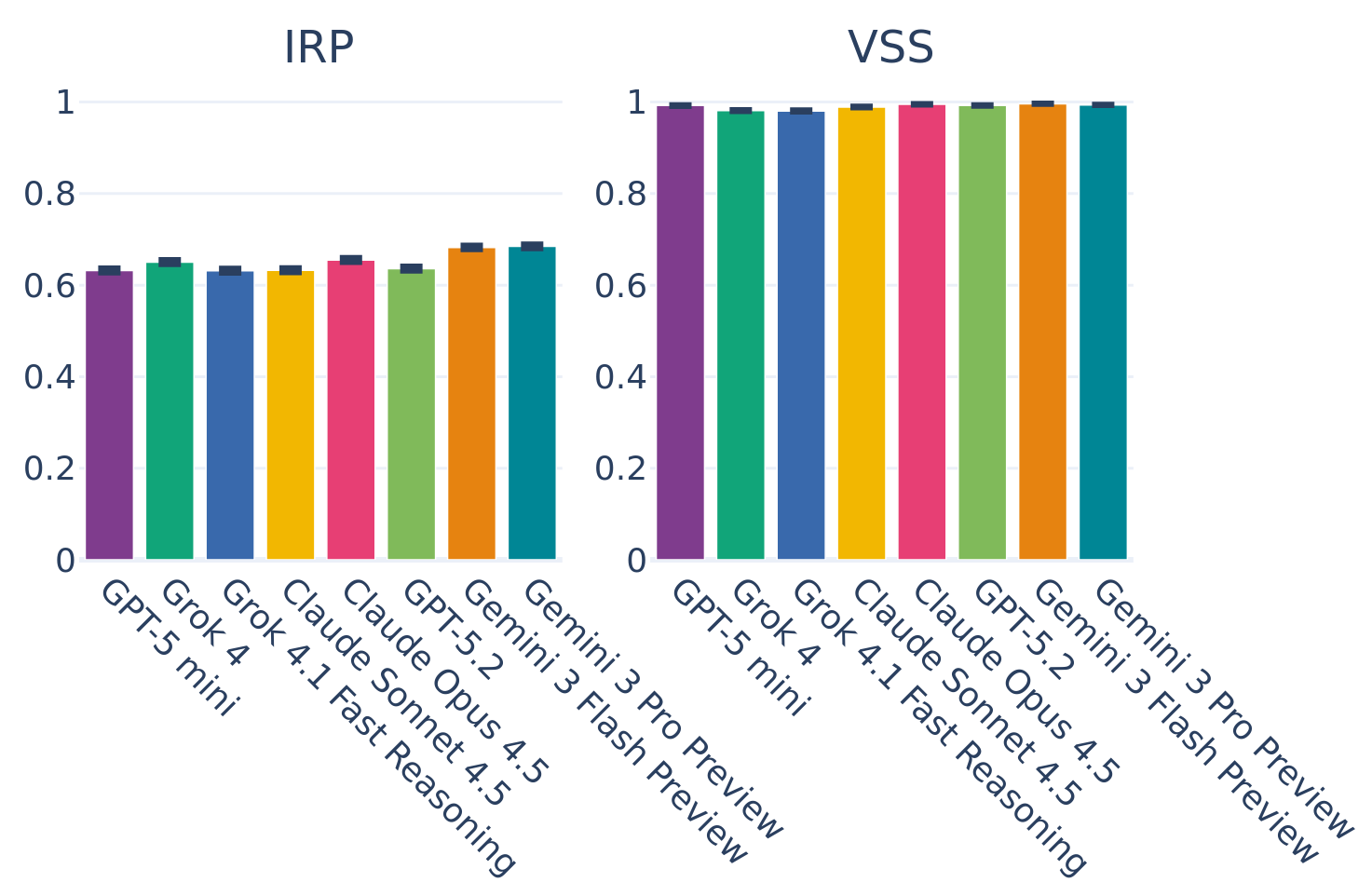}
        \caption{Identification Precision (IRP) and Voting Success Score (VSS)}
        \label{fig:metrics_voting}
    \end{subfigure}
    
    \vspace{0.5em}
    
    \begin{subfigure}[b]{0.6\textwidth}
        \centering
        \includegraphics[width=\textwidth]{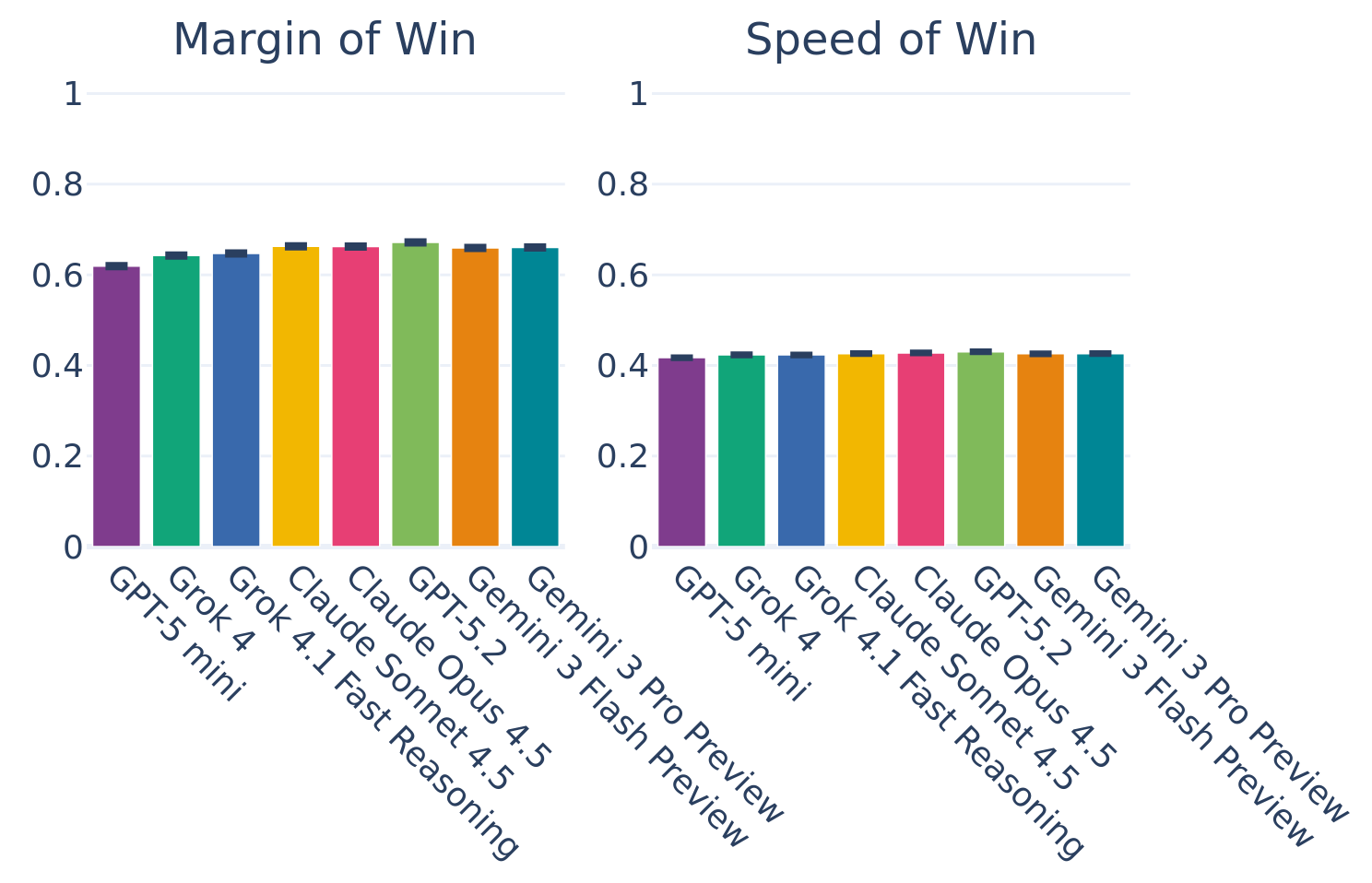}
        \caption{Margin of Win and Speed of Win}
        \label{fig:metrics_dominance}
    \end{subfigure}
\end{figure}

\clearpage 

\begin{figure}[htbp]
    \centering
    \addtocounter{figure}{-1} 
    \caption{Heuristic metrics across evaluated models, demonstrating the limited discriminative variance of heuristics versus ranking methods. The means and 95\% confidence intervals were derived from 1,000 bootstrap iterations across the 31,472 match-ups.}
    \label{fig:heuristic_metrics}
\end{figure}

Because these heuristics are heavily skewed by game structure and consensus incentives, they fail to isolate true capability. This reinforces our reliance on the Game Theoretic Evaluation (GTE) framework as the definitive, unbiased measure of model skill.


\section{Prompt Templates and Design Rationale}
\label{app:prompts}

All Game Arena environments share a common prompting philosophy: provide the model with a complete description of the game state, make the objective unambiguous, require step-by-step reasoning before a final answer, and keep everything else as minimal and neutral as possible so that observed behavior reflects the model's own capabilities rather than artifacts of prompt engineering. This appendix documents the exact templates, representative examples, and key design decisions for each game.

\subsection{Chess}
\label{app:chess_prompt}

The Chess prompt follows a shared template used across all Game Arena harnesses. At each turn, the model sees:

\begin{verbatim}
Let's play chess. The current game state in Forsyth-Edwards
Notation (FEN) notation is:
{fen}
The moves played so far are:
{pgn_movetext}
You are playing as player {color}.
It is now your turn. Play your strongest move. The move
MUST be legal. Reason step by step to come up with your
move, then output your final answer in the format
"Final Answer: X" where X is your chosen move in standard
algebraic notation (SAN).
\end{verbatim}

When a rethink is triggered---that is, the model's previous output was illegal or malformed---the harness appends a single line stating that the move was not legal and asking the model to try again. No further information is given: no explanation of \emph{why} the move was illegal, no indication of which pieces are pinned or attacked, and no list of legal alternatives.

A representative prompt delivered to a model playing Black in a late endgame is shown below:

\begin{verbatim}
Let's play chess. The current game state in Forsyth-Edwards
Notation (FEN) notation is:
8/8/8/1p3r2/8/5k1K/8/8 b - - 7 54
The moves played so far are:
1. e4 c5 2. Nf3 d6 3. d4 cxd4 4. Nxd4 Nf6 5. Nc3 a6
6. Bg5 e6 7. f4 Be7 8. Qf3 Qc7 9. O-O-O Nbd7 10. Bxf6
Nxf6 11. Nd5 exd5 12. exd5 Bg4 13. Qb3 Bxd1 14. Kxd1
O-O 15. Qc4 Qxc4 16. Bxc4 Ng4 17. Ke2 Bf6 18. c3
Rfe8+ 19. Kf3 Re3+ 20. Kxg4 Bxd4 21. cxd4 Re4 22. Rd1
h5+ 23. Kf5 Rae8 24. Bd3 Rxd4 25. Bc2 Rxd1 26. Bxd1
Re1 27. Bxh5 Rb1 28. b3 g6+ 29. Bxg6 fxg6+ 30. Kxg6
Rb2 31. f5 Rxg2+ 32. Kf6 Rxh2 33. Ke7 Rxa2 34. Kxd6
Kf7 35. Ke5 Re2+ 36. Kd6 Kf6 37. Kc7 Kxf5 38. Kb6
Re7 39. d6 Rd7 40. Kc5 Ke5 41. Kb6 Kxd6 42. b4 Rh7
43. b5 axb5 44. Kxb5 Kc7 45. Kc5 Rh5+ 46. Kd4 Kd6
47. Ke4 Kc5 48. Kf4 Kd4 49. Kg4 Re5 50. Kf4 b5
51. Kg4 Ke3 52. Kh3 Kf3 53. Kh4 Rf5 54. Kh3
You are playing as player Black.
It is now your turn. Play your strongest move. The move
MUST be legal. Reason step by step to come up with your
move, then output your final answer in the format
"Final Answer: X" where X is your chosen move in standard
algebraic notation (SAN).
\end{verbatim}

Several design choices merit explanation. The prompt includes both the FEN string and the full PGN movetext because some models---particularly those available when the benchmark was first introduced---struggled to parse FEN reliably but handled the movetext well. Including the move history also provides context for how the game has progressed: even though the optimal move depends only on the current position, the game history may help the model adapt to a specific opponent's style. The instruction to ``play your strongest move'' makes the optimization target unambiguous; without it, a model might reasonably interpret the task as playing a friendly or entertaining game. Finally, instructing the model to reason step-by-step before answering elicits chain-of-thought traces that tend to improve move quality and provide post-hoc interpretability.

\subsection{Poker}
\label{app:poker_prompt}

The poker harness communicates game state using a format adapted from the PokerStars hand history standard. This widely recognized convention is unambiguous, information-complete, and likely well-represented in model training data, ensuring that formatting artifacts do not confound the evaluation of strategic reasoning. Within each 100-hand episode, the model receives the cumulative history of all prior hands followed by the current hand in progress, enabling within-episode opponent modeling.

A complete hand as presented to the model is shown below:

\begin{verbatim}
Hand #13: Hold'em No Limit (1/2)
Table '' 2-max (USD) Seat #1 is the button
Seat 1: GPT-5 mini (200 in chips)
Seat 2: Claude Sonnet 4.5 (200 in chips)
GPT-5 mini: posts small blind 1
Claude Sonnet 4.5: posts big blind 2
*** HOLE CARDS ***
Dealt to GPT-5 mini [Tc Th]
Dealt to Claude Sonnet 4.5 [4c Qc]
GPT-5 mini: raises 4 to 6
Claude Sonnet 4.5: calls 4
*** FLOP *** [Qs 7d 6s]
Claude Sonnet 4.5: checks
GPT-5 mini: bets 4
Claude Sonnet 4.5: calls 4
*** TURN *** [Qs 7d 6s] [Ad]
Claude Sonnet 4.5: checks
GPT-5 mini: checks
*** RIVER *** [Qs 7d 6s] [Ad] [4s]
Claude Sonnet 4.5: bets 10
GPT-5 mini: calls 10
*** SHOWDOWN ***
Claude Sonnet 4.5: shows [4c Qc]
Claude Sonnet 4.5 collected 40.0 from pot
*** SUMMARY ***
Total pot 40 | Rake 0
Board [Qs 7d 6s Ad 4s]
Seat 1: GPT-5 mini (button) (small blind) showed [Tc Th]
  and lost
Seat 2: Claude Sonnet 4.5 (big blind) showed [4c Qc]
  and won (40.0)
\end{verbatim}

In addition to the hand history, the model receives a system prompt developed in consultation with poker professionals. It instructs the model to maximize expected value using game-theory-optimal play as a baseline, to deviate from GTO when it identifies exploitable tendencies in the opponent's observed play, and to produce an explicit reasoning trace grounded in core poker concepts (range advantage, pot odds, fold equity, position) before declaring an action. This framing was chosen to eliminate ambiguity about the objective (EV maximization, not entertainment or risk-averse survival) and to produce interpretable traces for post-hoc analysis---pilot experiments showed that without the reasoning-trace instruction, models frequently omit strategic justification entirely.

As in Chess, the harness does not enumerate legal actions; the model must infer from the game state what options are available (fold, check, call, raise, and the range of legal bet sizes). Providing the full within-episode hand history allows models to identify and exploit opponent patterns (e.g., excessive folding to 3-bets, or predictable sizing), a core component of expert-level heads-up play. The 100-hand episode length balances this adaptability against context-window constraints.

\subsection{Werewolf}
\label{app:Werewolf_prompt}

The Werewolf harness uses a ReAct-style prompting framework in which, at each decision point, the model receives a composite prompt assembled from four components: (1)~a system prompt defining the game rules, the model's assigned role, and the team-victory objective---explicitly targeting \emph{team} victory rather than individual survival, a distinction that matters strategically because, e.g., a Villager may need to sacrifice themselves to expose a Werewolf; (2)~the current public and role-specific private state (who is alive, prior vote outcomes, the Seer's investigation results, etc.); (3)~a chronological memory log of all observed events and the model's own prior reasoning, accumulated across the game; and (4)~a phase-specific task instruction appropriate to the current game phase (e.g., ``Submit your vote for elimination,'' ``Write a message to the group,'' or ``Choose a player to investigate'').

The model returns a JSON object with two fields: a private \texttt{reasoning} field (not visible to other players, used for analysis) and a public \texttt{action} field (the chat message, vote, or role ability to be executed). This separation ensures that internal deliberation does not inadvertently leak information. Output reliability is managed through lenient JSON parsing (\texttt{pyjson5}), exponential-backoff retries with intelligent context truncation on overflow, and a rule-based safety filter for violent or inappropriate language.

\section{Authorship by Area}

The \textbf{Game Arena platform and benchmark.} was developed as a collaboration between Google DeepMind (GDM), Kaggle, and the Google Cloud Office of the CTO (OCTO).

\paragraph{GDM Tournament Infrastructure \& OpenSpiel Integration.}

John Schultz and Justin Chiu built the foundational tournament infrastructure and OpenSpiel integration for Chess and Poker, with contributions to design and operations from Clayton Drazner, Marc Lanctot, Phoebe Kirk, Karim Hakimzadeh, Yi Su and Minmin Chen.

\paragraph{Werewolf Game Development.}

The Werewolf tournament was conceived and built by Hann Wang, Chuck Sugnet, and Tom Mason under the guidance of Diane Chaleff and Antonio Gulli (OCTO).

\paragraph{Evaluation Methodology \& Metrics.}

The evaluation harness and game-specific protocols were developed by Timothy Chung, John Schultz, Ian Gemp, Siqi Liu, Elsa Dong, Dima Yeroshenko and Minmin Chen. Yuchen Zhuang, Jie Ren and Andrew Lee and Sahand Sharifzadeh researched majority voting, retry logic, and sequential revision strategies. Key contributions to leaderboarding and metrics were made by Daniel Hennes, John Schultz, Timothy Chung, Yuexiang Zhai, Bo Chang, Kate Larson, Nenad Tomasev, Ya Xu, Martyna Plomecka, Yao Yan, and Will Cukierski.

\paragraph{Visualizers \& UI Design.}

Game visualizers and the streaming UI were designed and built by Yuting Han, Christopher D'Mello, Chris Prichard, Domino Weir, Michael Aaron, Riley Jones, and Johnny Yip.

\paragraph{Kaggle Platform \& Infrastructure Engineering.}

The Kaggle platform infrastructure for Game Arena was built by Bovard Doerschuk-Tiberi, Jon Lipovetz, Andrew Wang, Jeff Moser, Chiamaka Chukwuka, DJ Sterling, and Jun Peng. This platform was used to produce all of the results in this work.

\paragraph{Research Analysis \& Publication.}

The research analysis pipeline and paper authorship were co-led by Martyna Plomecka and Yao Yan, with contributions from John Schultz, Hann Wang, Timothy Chung, Will Cukierski and Minmin Chen. 

\paragraph{Operations \& Program Management.}

Program management was led by Laurel Prince and Addison Howard. Meghan O'Connell managed executive coordination. 

\paragraph{Project Leadership, Vision \& Strategy.}

The game arena was envisioned in a discussion between Demis Hassabis, Minmin Chen, Ya Xu, Kate Olszewska and Nate Keating.

PM leads: Kate Olszewska, Nate Keating, Meg Risdal, Oran Kelly

Research leads: Minmin Chen, Orhan Firat

Engineering leads: Robert Fraser

Strategic Advisory :Rupert Kemp and Saaber Fatehi 

\paragraph{Go-to-Market, Communications \& Partnerships.}

GTM strategy and execution were run by Kinjal Parekh, Jaimie Hwang, Praveen Kumar Rajasekar, Melissa Nalubwama, Harrison Jobe, Ryan Trostle, Lloyd Hightower, and Roxanne Daniel. Chad Woodford provided legal review for games and data releases.

\end{document}